\documentclass[10pt,twocolumn,letterpaper]{article}

\usepackage[pagenumbers]{cvpr} % To force page numbers, e.g. for an arXiv version

\usepackage{graphicx}
\usepackage{amsmath}
\usepackage{amssymb}
\usepackage{booktabs}
\usepackage{color}
\usepackage[numbers,sort]{natbib}
\usepackage{multirow}
\usepackage{array}
\usepackage{verbatim}
\definecolor{hollywoodcerise}{rgb}{0.96, 0.0, 0.63}
\definecolor{lasallegreen}{rgb}{0.03, 0.47, 0.19}
\definecolor{hanpurple}{rgb}{0.32, 0.09, 0.98}
\definecolor{green(pigment)}{rgb}{0.0, 0.65, 0.31}
\usepackage[pagebackref=true,breaklinks=true,letterpaper=true,colorlinks,bookmarks=false]{hyperref}

\usepackage[capitalize]{cleveref}
\crefname{section}{Sec.}{Secs.}
\Crefname{section}{Section}{Sections}
\Crefname{table}{Table}{Tables}
\crefname{table}{Tab.}{Tabs.}

\def\cvprPaperID{****} % *** Enter the CVPR Paper ID here
\def\confName{CVPR}
\def\confYear{2022}

\begin{document}

% \title{SphereSR: 360$^{\circ}$ Panoramic Image Super-Resolution for Arbitrary Field-of-View and Projection-Type using Continuous Image Representation}

\title{SphereSR: 360$^{\circ}$ Image Super-Resolution with Arbitrary Projection via Continuous Spherical Image Representation}

\author{Youngho Yoon, Inchul Chung, Lin Wang, and Kuk-Jin Yoon\\
Visual Intelligence Lab., KAIST, Korea\\
{\tt\small \{dudgh1732,inchul1221,wanglin,kjyoon\}@kaist.ac.kr }
%\and
}
\maketitle

%%%%%%%%% ABSTRACT
\begin{abstract}

The 360$^{\circ}$ imaging has recently gained great attention; however, its angular resolution is relatively lower than that of a narrow field-of-view (FOV) perspective image as it is captured by using fisheye lenses with the same sensor size.
% in many fields. %owing to its remarkably large flied-of-view (FOV).
% However, 
% % as a 360$^{\circ}$ image is , 
% the angular resolution of a 360$^{\circ}$ image is relatively lower than that of a narrow flied-of-view (FOV) perspective image as it is captured by using fisheye lenses with the same sensor size.
%however, most 360$^{\circ}$ images are usually at a low resolution (LR) due to the limitation of sensors, storage, etc. 
Therefore, it is beneficial to super-resolve a 360$^{\circ}$ image.
% with better quality.
% in many cases. 
%Some attempts have been made to achieve this goal, and most of them focused the equirectangular projection (ERP) type although there are various ways to project a 360$^{\circ}$ image. 
Some attempts have been made but mostly considered the equirectangular projection (ERP) as one of the way for 360$^{\circ}$ image representation despite of latitude-dependent distortions.
% among various ways of the projection types,  
% among various ways of 360$^{\circ}$ image 
% projection for the 360$^{\circ}$ image representation, although there exist latitude-dependent distortions. 
In that case, as the output high-resolution (HR) image is always in the same ERP format as the low-resolution (LR) input, another information loss may occur when transforming the HR image to other projection types. 
% of 
% 360$^{\circ}$ image 
% projection. 
%  Recently, as the use of 360-degree images has been increased, network development for Panoramic Image Super-Resolution is being actively carried out. 
%  however, despite the characteristics of a panoramic image that can be converted through various projection types, previous works were conducted on the ERP projection type, which is one of the panoramic image projection types. 
%  In addition, . 
% To solve this problem,
%In this paper, we propose a novel framework to generate a continuous spherical image representation, called SphereSR, \textbf{aiming to learn from an LR 360$^{\circ}$ image and predict the RGB values of spherical coordinate of an HR image with an arbitrary projection type}. 
In this paper, we propose SphereSR, a novel framework to generate a continuous spherical image representation from an LR 360$^{\circ}$ image, aiming at predicting the RGB values at given spherical coordinates for super-resolution with an arbitrary 360$^{\circ}$ image projection. 
% SphereSR consists of two parts. 
%Importantly, we first propose a spherical data structure based on icosahedron and convolution kernel method to efficiently extract spherical features on the surface composed of uniform meshes. 
Specifically, we first propose a feature extraction module that represents 
the spherical data based on icosahedron and efficiently extracts features on the spherical surface. 
% composed of uniform meshes.
We then propose a spherical local implicit image function (SLIIF) to predict RGB values at the spherical coordinates. As such, SphereSR flexibly reconstructs an HR image under an arbitrary projection type.
% sphere-oriented cell decoding that generalizes to an arbitrary projection type. 
% Lastly, we propose a mesh-based local implicit image function to extract RGB values for arbitrary spherical coordinate points. 
% In particular first propose a spherical data structure and convolution kernel method. 
% We then propose a mesh-based local implicit image function to extract RGB values for arbitrary spherical coordinate points. 
% Then, we propose a mesh-based local implicit image function that can extract RGB values for arbitrary spherical coordinate points. 
Experiments on various benchmark datasets show that our method significantly surpasses existing methods.

%In this paper, we propose a continuous spherical image representation method, called SphereSR, aiming to super-resolve an LR 360$^{\circ}$ to an HR image with an arbitrary projection type. In particular, to generate an HR image with an arbitrary projection type, we propose to a spherical kernel to efficiently represent the 360 image and extract spherical feature maps from an encoder and generate an HR image oriented to the grid through the sphere-oriented cell decoding. Moreover, as directly learning a continuous representation for an LR 360$^{\circ}$ image takes considerable memory load, we thus propose a spherical data structure to the spherical images. Experiments on various benchmark datasets show that our method significantly surpasses existing methods.

%  In this paper, we introduce a spherical continuous image representation method capable of super-resolving an LR 360$^{\circ}$ image with an arbitrary projection-type. 
%  To generate a HR output with an arbitrary projection-type, we first a spherical feature map is first extracted, and then generate an HR image oriented to the grid through the sphere-oriented cell decoding. 
%  In addition, we propose a novel spherical data structure and sphere kernel
% %  that can solve the memory issue of spherePHD[] 
%  to propose a suitable network for spherical image super-resolution.
%  [also describe experiments]
\end{abstract}

%%%%%%%%% BODY TEXT
\vspace{-15pt}
\section{Introduction}
\vspace{-5pt}
\label{sec:intro}

\begin{figure}[t!]
  \centering
  \includegraphics[width=0.93\linewidth]{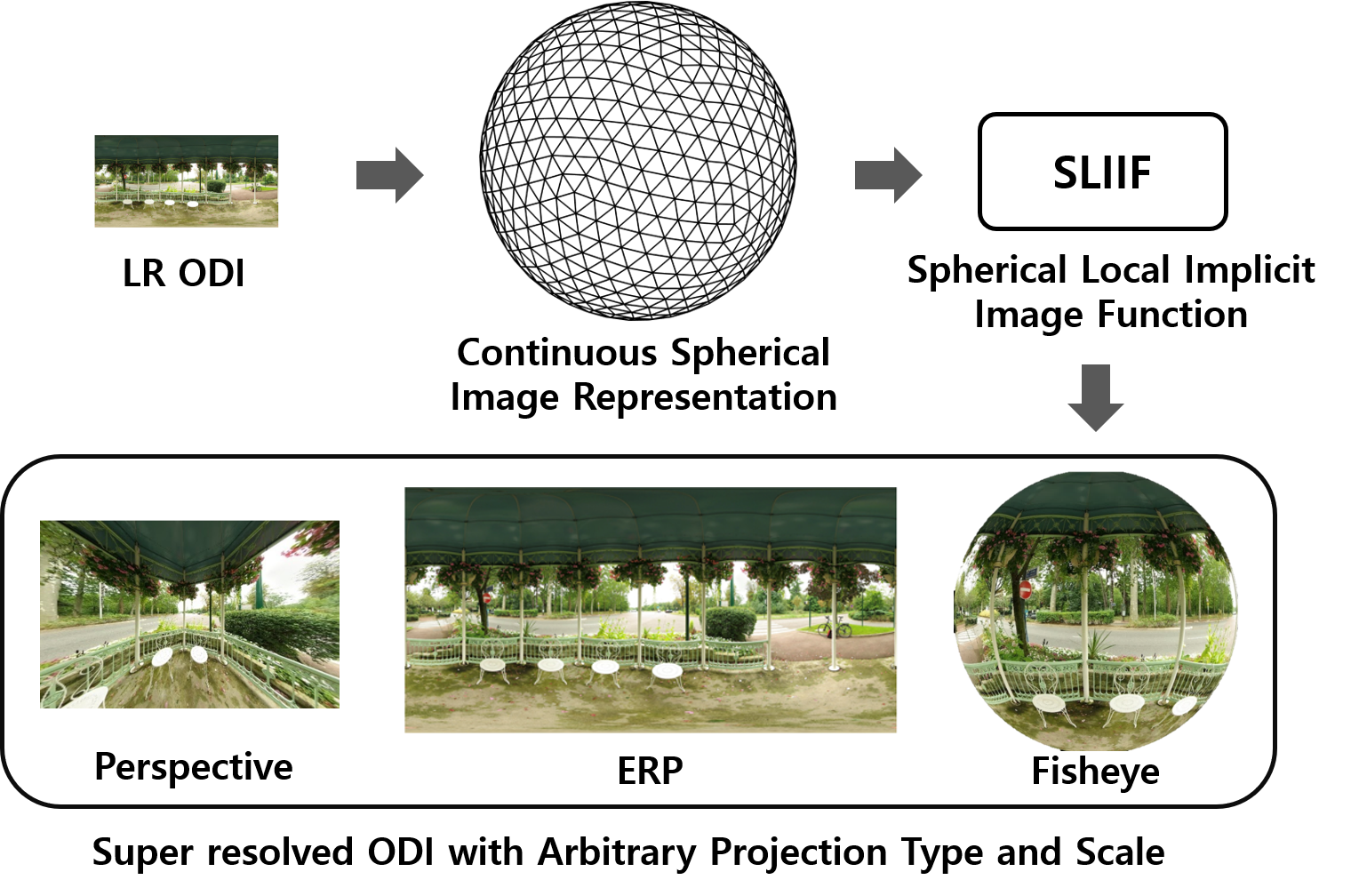}
   \vspace{-10pt}
   \caption{Learning continuous spherical image representation. SphereSR leverages SLIIF to predict the RGB values at given spherical coordinates for SR with arbitrary image projection.}
   \label{fig:first_page}
   \vspace{-10pt}
\end{figure}

The 360$^{\circ}$ imaging has recently gained great attention in many fields including AR/VR. %attention owing to its higher flied-of-view (FoV) than the normal FoV image. 
In general, raw 360$^{\circ}$ images are transformed into 2D planar representations preserving the omnidirectional information, \eg, equirectangular projection (ERP)~\cite{kim2020super} and cube map projection (CP)~\cite{kim2020super} to be compatible with the
% off-the-shelf 
imaging pipelines. The omnidirectional images (ODIs)\footnote{Throughout the paper, we use omnidirectional images and 360$^{\circ}$ images interchangeably.} are sometimes projected back onto a sphere or transformed with different types of projection and rendered for display in some applications. 
% it always provides a perspective view for any direction, 360$^{\circ}$ imaging is widely explored in autonomous driving \cite{}, VR~\cite{}, and street view map~\cite{}.

However, the angular resolution of a 360$^{\circ}$ image tends to be lower than that of a narrow field-of-view (FOV) perspective image as it is captured using fisheye lenses with the same sensor size. Moreover, the 360$^{\circ}$ image quality can be degraded during the transformation between different image projection types. 
% the ODI representation introduces some critical challenges as it requires exceptionally high-resolution (HR) to cover the whole 360$^{\circ}$ scene.
Therefore, it is imperative to super-resolve the low-resolution (LR) 360$^{\circ}$ image by considering various projections to provide a high-level visual quality in diverse conditions. 
Early studies have attempted to reconstruct high-resolution (HR) ODIs by interpolating the missing data between the LR image pixels~\cite{nagahara2000super,arican2011joint,bagnato2010plenoptic}. 

Recently, deep learning (DL) has brought a significant performance boost for the 2D single image super-resolution (SISR)~\cite{zhang2018image, lai2017deep,tai2017Memnet}. These methods mostly explore super-resolving the 2D LR image using high-capacity convolutional neural networks (CNNs), via \eg, residual connections~\cite{lim2017enhanced}, and learning algorithms including generative adversarial networks (GANs)~\cite{wang2018esrgan,ledig2017photo,wang2022semi}. 
However, directly using these methods for 360$^{\circ}$ images represented in 2D planar representations is less applicable as the pixel density and texture complexity vary across different positions in 2D planar representations of 360$^{\circ}$ images, as pointed in \cite{deng2021lau}.

Consequently, some attempts have been made to address the SR problems for 360$^{\circ}$ imaging~\cite{ozcinar2019super,deng2021lau,Sun2017WeightedtoSphericallyUniformQE,Zhou2018WeightedtoSphericallyUniformSO}. In particular, 360-SS~\cite{ozcinar2019super} proposes a GAN-based framework using the Pix2Pix pipeline~\cite{isola2017image}; however, it focuses only on the ERP format and does not fully consider the properties of 360$^{\circ}$ images. 
LAU-Net~\cite{deng2021lau} introduces a method to identify distortions of an ODI on the latitude and upsample the ODI pixels on the segmented patches. However, it leads to considerable disconnections along the patches.
% caused by the less effective segmentation threshold. %, as shown in Fig.~\ref{fig:first_page}(c). 
% Thus, the ERP image distortion problem cannot be directly solved. 
% the connection between the patches split for each latitude in the middle of the network is cut off, and the image distortion phenomenon cannot be directly solved, as shown in Fig.~\ref{fig:onecol}(c).
% the characteristics of ODIs that are distorted according to latitude were identified, and the performance of SR was improved by designing a model optimized for these characteristics. However, this paper has the point that the connection between the patches split for each latitude in the middle of the network is cut off, and the image distortion phenomenon cannot be directly solved.
In a nutshell, existing methods for ODI SR ignore the projection process of 360$^{\circ}$ images in real applications and only take the ERP image as the LR input and produce the HR ERP output. %Moreover, projecting an HR ERP image to a sphere and rendering it for display needs interpolation operations, thus leading to blurry results. 
%
% In this case, since the SR image selected at HR is interpolated with a specific projection type again, a blurry result is inevitable.
%
Indeed, the 360$^{\circ}$ image can be flexibly
% has the property of being able to be 
converted into various projection types as, in real applications, user specifies the projection type, direction, and FOV. Thus, it is vital to address the ERP distortions problems and strive to super-resolve an ODI image to an HR image with \emph{an arbitrary projection type} rather than a fixed type. 

% Inspired by \cite{}
% multi-modal learning~\cite{} and continuous image representation methods~\cite{}, 
In this paper, as shown in Fig.~\ref{fig:first_page}, we propose a novel framework, called SphereSR, aiming at super-resolving an LR 360$^{\circ}$ image to an HR image \emph{with an arbitrary projection type via continuous spherical image representation}. 
% SphereSR consists of three key methods.
First, 
% to achieve a better spherical representation, 
we propose a feature extraction module that represents the spherical data based on icosahedron and efficiently extracts features on the spherical surface composed of uniform faces (Sec.~\ref{ds_ks}). As such, we solve the ERP image distortion problem and the pixel density difference according to the latitude.
% mentioned in LAU-Net at once 
% by introducing a convolution method on the meshes of the sphere created through subdivision of icosahedron.
Second, we propose a spherical local implicit image function (SLIIF) that can predict RGB values at arbitrary coordinates of a sphere feature map, inspired by LIIF~\cite{chen2021learning} (Sec.~\ref{image_func}). 
SLIIF works on \emph{triangular face}, buttressed by the position embedding based on the normal plane polar coordinates to obtain relative coordinates on a sphere. Therefore, our method tackles pixel-misalignment issue when being projected to another ODI projection.  
% Lastly, we propose a 
% % sphere-oriented 
% cell decoding method that calculates the pixel area between a sphere and an output projected by an arbitrary projection type (Sec.~\ref{cell_decode}).  
% Therefore, it makes  the  network  train  a  relationship  between  a  pixel’s size and a SR image RGB values.
As a result, SphereSR can predict RGB values for any SR scale parameters. Additionally, to train SphereSR, we introduce a feature loss that measures the similarity between two projection types, leading to considerable performance enhancement (Sec.~\ref{loss_func}).
% by calculating cell decoding values for each pixel.
Extensive experiments on various benchmark datasets show that our method significantly surpasses existing methods.
% The distortion problem is directly solved by convolution on the sphere, and SR results for various projection types can be extracted by introducing a continuous image representation method.

In summary, the contributions of our paper are four-fold. (I) We propose a novel framework, called SphereSR, aiming at super-resolving an LR 360$^{\circ}$ image to an HR image with an arbitrary projection type.
(II) We propose a feature extraction module that represents the spherical data based on icosahedron and extracts features on spherical surface. (III) We propose SLIIF that predicts RGB values from the spherical coordinates. (IV) Our method achieves the significantly better performance in the extensive experiments. 

\section{Related Works}
\noindent \textbf{Omnidirectional Image SR and Enhancement.} % Omni-Directional Image Super-Resolution
% As the importance and popularity of ODI increase, there have been various attempts for enhancing ODI to overcome the limitation of low resolution and quality. 
Early ODI SR methods~\cite{Nagahara2000SuperresolutionFA, Kawasaki2006SuperresolutionOC, Arican2009L1RS, Bagnato2010PlenopticBS, Arican2011JointRA} focused on assembling and optimizing multiple LR ODIs 
% processes were conducted
on spherical or hyperbolic surface.
% which are original shape of ODI images.
On the other hand, as the distortion in ODI is caused due to the projection of original spherical image to the 2D planar image plane, recent research focused on tackling and solving distortion in ODI using 2D convolution to achieve qualitative result in observation space, \ie spherical surface. 
%Recently, research has been focused on tackling the solving distortion in ODI using 2D convolution.
% various researches have been proposed owing advantage of 2D convolution. 
% To handle aspects different from ordinary 2D image enhancement, 
% most of proposed methods focus on solving distortion in ODI. 
Su \etal~\cite{Sun2017WeightedtoSphericallyUniformQE} and Zhou \etal ~\cite{Zhou2018WeightedtoSphericallyUniformSO} proposed evaluation methods for ODI weighted with the projected area on the spherical surface. 
%\cite{Nishiyama2021360SI} adapted existing SISR model by adding distortion map as input to super-resolve ERP images by considering different distortion. 
\cite{Sevom2018360PS} and \cite{Nishiyama2021360SI} adapted existing SISR models to ERP SR by fine-tuning or by adding a distortion map as an input to tackle different distortions.
\cite{ozcinar2019super} leveraged GAN to super-resolve an ODI by applying WS-SSIM~\cite{Zhou2018WeightedtoSphericallyUniformSO}. \cite{Zhang2020TowardRP} also proposed the GAN-based framework employing multi-frequency structures to enhance the panoramic image quality up to the high-end camera quality. \cite{Liu2020ASF} focused on the 360$^{\circ}$ image SR utilizing single-frame and multi-frame joint learning and a loss function weighted differently along the latitude. \cite{deng2021lau} considered varying pixel density and texture complexity along latitude by proposing network allowing distinct up-scaling factors along the latitude bands.
\emph{Unlike the aforementioned methods, we propose to predict the RGB values at the given spherical coordinates of an HR image with respect to an arbitrary project type from an LR 360$^{\circ}$ image.}
% method performing super-resolution task on original spherical surface resulting in minimization of the distortion among image. 

%and maintain uniformity over any direction.
%However our model performs convolution on sperical feature map resulting to no need of considering distorsion and imbalance along image.

% \noindent \textbf{Deep learning-based 2D SISR} %Arbitary scale SISR
\noindent \textbf{2D SISR with an Arbitrary Scale.}
% SISR has led to rapid performance improvement through various deep-learning based methods.[] 
Research on SISR with an arbitrary scale has been actively conducted. \cite{lim2017enhanced} first proposed a method that enables multiple scale factors over one network. MetaSR~\cite{hu2019meta} achieves SR with the non-integer scale factors. However, both methods are limited to SR with the symmetric scales. Later on, \cite{wang2020learning} proposed a framework that enables asymmetric 
% SR with different 
scale factors along horizontal and vertical axes. Moreover, SRWarp~\cite{son2021srwarp} generalizes SR toward an arbitrary image transformation. 
% possible in parity image transformation using a differentiable warping module.
Although these methods are effective for 2D SISR with an arbitrary scale factor, they fail to be directly applied to 360$^{\circ}$ image SR
% there is still a limit to the immediate use of existing research
due to the difference between xy-coordinate (2D) and spherical coordinates in ODI domains. \emph{We overcome the challenge by proposing SphereSR that leverages SLIIF to predict RGB values for the arbitrary spherical coordinates}. 

\begin{figure*}[t!]
  \centering
  \includegraphics[width=.85\linewidth]{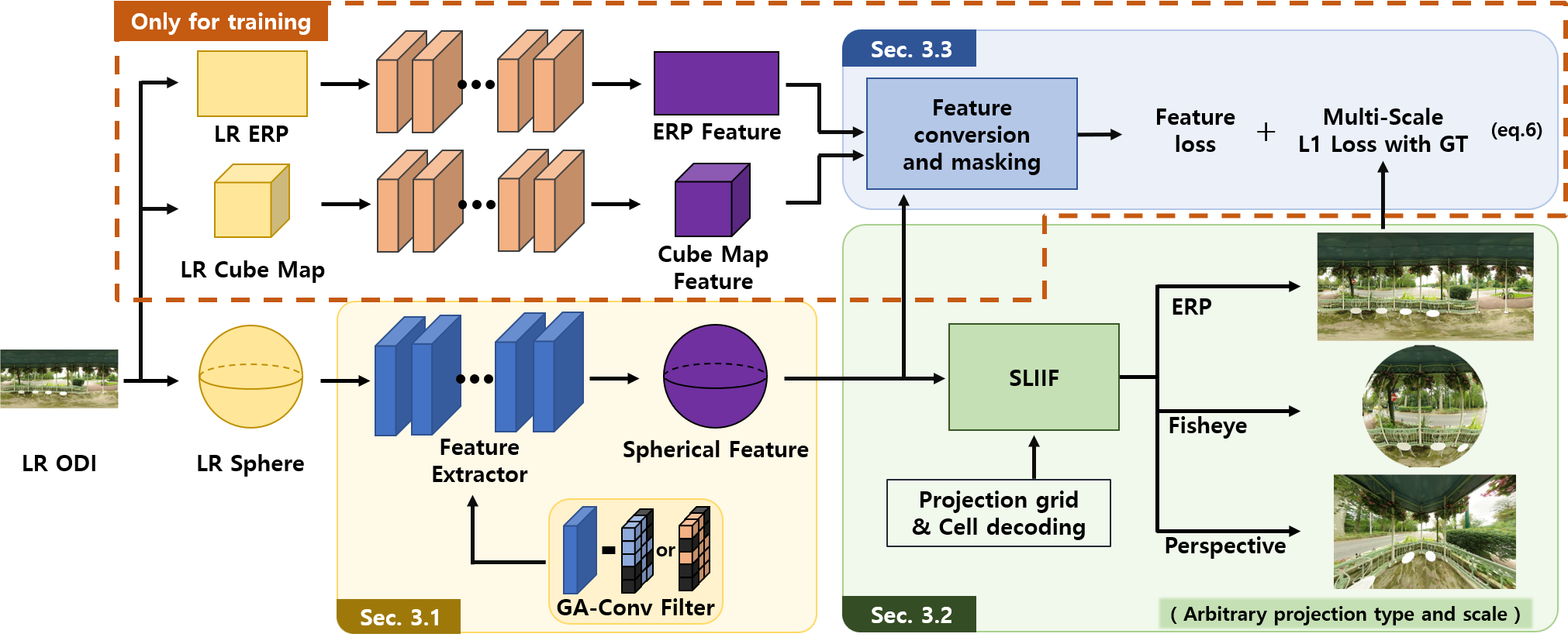}
   \vspace{-5pt}
   \caption{Overall framework of the proposed SphereSR.}
   \label{fig:overall-framework}
   \vspace{-10pt}
\end{figure*}

\noindent\textbf{Continuous Image Representation.}
Research on implicit neural representation (INR) has been conducted to express 3D spaces, \eg, 3D shape construction and novel view synthesis as continuous ways~\cite{mildenhall2020nerf,sitzmann2019scene,mescheder2019occupancy}. Since then, continuous image representation has been explored on the (x,y) coordinate. \cite{anokhin2021image,skorokhodov2021adversarial} used the networks to predict the RGB value of each pixel from the latent vector on (x,y) coordinate without a spatial convolution for 2D image generation. LIIF~\cite{chen2021learning} proposes to bridge between discrete and continuous representation for images on the (x,y) coordinate.
%We introduce continuous spherical image presentation to extend to $(\theta,\phi)$ coordinate on the unit sphere. 
\emph{We propose SLIIF that makes continuous image representation on the unit sphere possible.}

\noindent \textbf{CNNs for Spherical Images.}
% Many studies have proposed various CNN architectures with the ODI input. 
\cite{cohen2018spherical} proposed a CNN-based method on the sphere with structural characteristics called rotational equivariance. However, it requires Fourier transform for each step.
% making it difficult to apply existing methods. 
% is very different from the existing CNNs methodology, 
% making it difficult to develop their model through existing studies.
\cite{coors2018spherenet} developed a CNN filter in light of spatial location on the sphere to solve the distortion problem of the ERP images. \cite{su2019kernel} proposed a kernel transformer network that converts the pre-trained kernels on the perspective images into ODIs.
% \cite{eder2020tangent} 
% conducted convolution through the process of generating spatial images into several sets of local-planar images using inverse genomic projects.
% addressed data orientation by ensuring all tangent images are consistently oriented when rendering and circumvents the discretization issue by rendering to image pixel grids.
SpherePHD~\cite{lee2020spherephd} proposed a convolution kernel applicable to triangular pixels defined on the faces of the icosahedron.
% representation through the convolution kernel shape applicable to all pixels from the triangular faces subdivision of the icosahedron. 
\cite{zhang2019orientation} performed convolution using a hexagonal filter applicable to the vertices of icosahedron. 
In this work, we focus on ODI SR and propose SphereSR that applies convolution to the spherical structure created through the subdivision of the icosahedron. 
% \emph{In particular, we propose a feature extraction module applied on faces of icosahedron enabling efficient spherical feature extraction.}

% are proposed by acquiring a SpherePHD method 
%predicts RGB values of arbitrary points on triangular pixels based on convolution operation on faces.}

%on the triangular faces for image representation on uniform icosaherdon mesh.}

%%%%%%%%%%%%%%%%%% Method %%%%%%%%%%%%%%%%%%
\begin{figure}[t!]
  \centering
  \vspace{-5pt}
  \includegraphics[width=.98\linewidth]{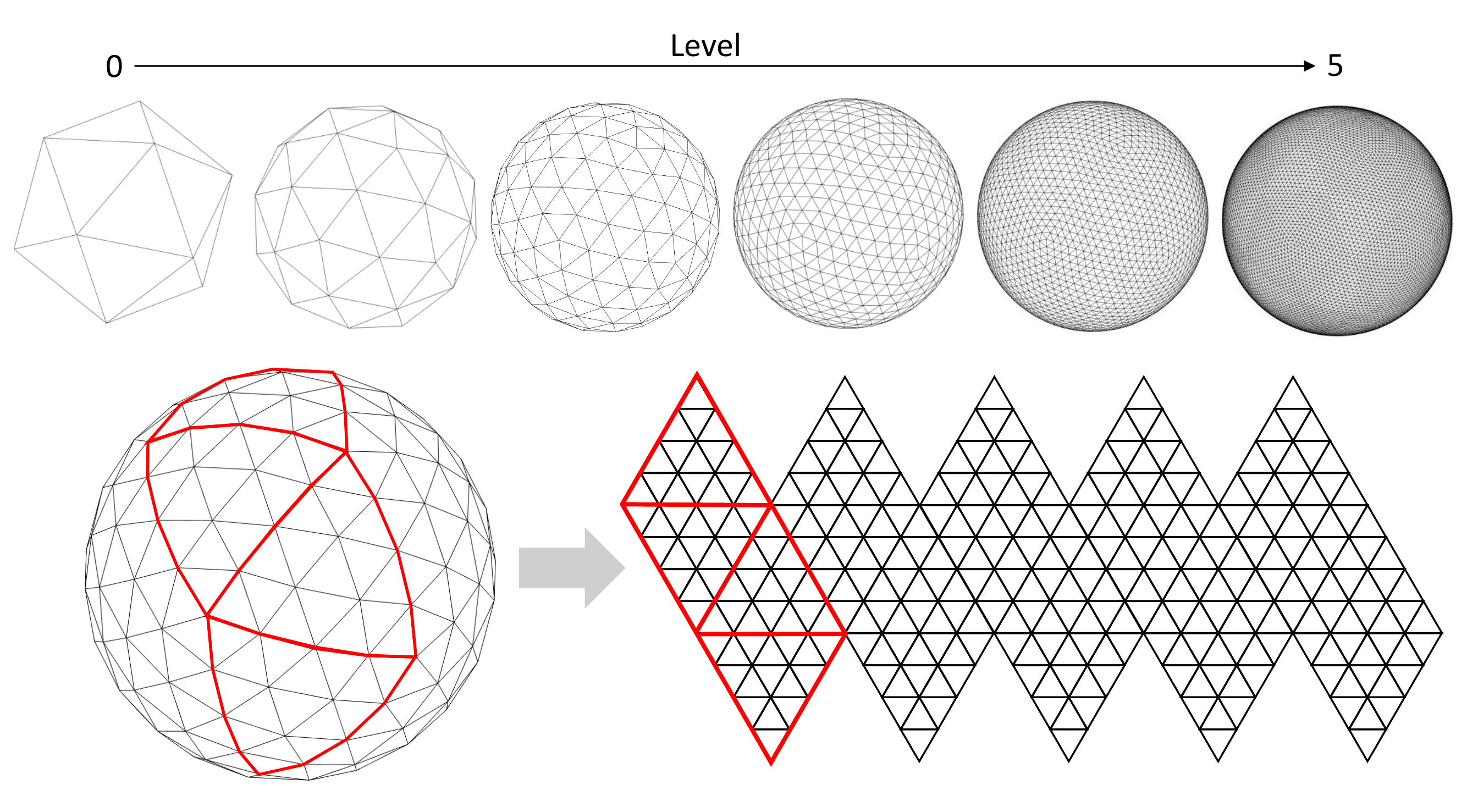}
   \vspace{-6pt}
   \caption{Subdivision process of icosahedron. We define a pixel as a face of subdivided icosahedron.}
   \label{fig:icosahedron}
   \vspace{-12pt}
\end{figure}
%overview-new spherical CNN methods / kernel weight sharing assignment
%Spherical local implicit Image function - feature unfolding/local ensemble/cell decoding
% 3.1 overview
% overall framework
% 우리의 방법은 panorama image의 original shape이라고 볼 수 있는 sphere상에서 panorama image super-resolution을 수행하는 것이다. spherical input에 대해서 super-resolution을 진행하기 위해서는 먼저 spherical meshes에 대한 convolution method를 필요로 하고, 두 번째로 spherical feature representation을 임의의 projection type으로 변환할 수 있는 mlp module을 필요로 한다.

% 3.2 local implicit image function on spherical coordinate

% 3.3 sphere convolution by new data structure / weight sharing method

%\vspace{-6pt}
% \section{Method}
 \vspace{-3pt}
\section{Method}
 \vspace{-3pt}
\noindent\textbf{Overview.} 
As shown in Fig.~\ref{fig:overall-framework},
%, we generate an output with arbitrary projection type through the proposed continuous representation. To this end, 
we propose a novel framework, SphereSR, that aims at obtaining a continuous spherical image representation from a given icosahedron input. % and generate output with arbitrary projection type.
First, we introduce a feature extraction method for spherical images that efficiently extracts features from an image on the icosahedron (Sec.~\ref{ds_ks}). 
Second, we propose the Spherical Local Implicit Image Function (SLIIF) that can predict RGB values through the extracted features in order to flexibly reconstruct an HR image with arbitrary projection type(Sec.~\ref{image_func}). 
%마지막으로 우리는 여러가지 projection type으로 변환가능한 SLIIF의 장점을 활용하여 여러 가지 프로젝션 인풋으로부터 생성된 features로부터 feature similarity를 측정할 수 있는 feature similarity loss를 제안한다.
Lastly, We propose a feature loss to obtain support from features of other projection types by utilizing the advantage of SLIIF that can be converted to arbitrary projection type(Sec.~\ref{loss_func}).
\subsection{Feature Extraction for Spherical Images}
\label{ds_ks}
% \subsection{crystal Convolutions}
% \subsection{DIA(dinamic-aligned) Convolutions}

Feature extraction is crucial yet challenging for spherical image SR as we focus on very large scale factors, \eg, $\times$16. In this situation, \emph{it is imperative to tackle the memory overload issue while ensuring high SR performance}. For that reason, the proposed SphereSR represents the spherical data based on icosahedron and efficiently extracts features on the spherical surface composed of uniform faces. This is achieved by a new data structure on icosahedron and weight sharing between kernels of different directions. %We now present the technical details of them.

\begin{figure}[t]
 \vspace{-5pt}
 \centering
  \includegraphics[width=.98\linewidth]{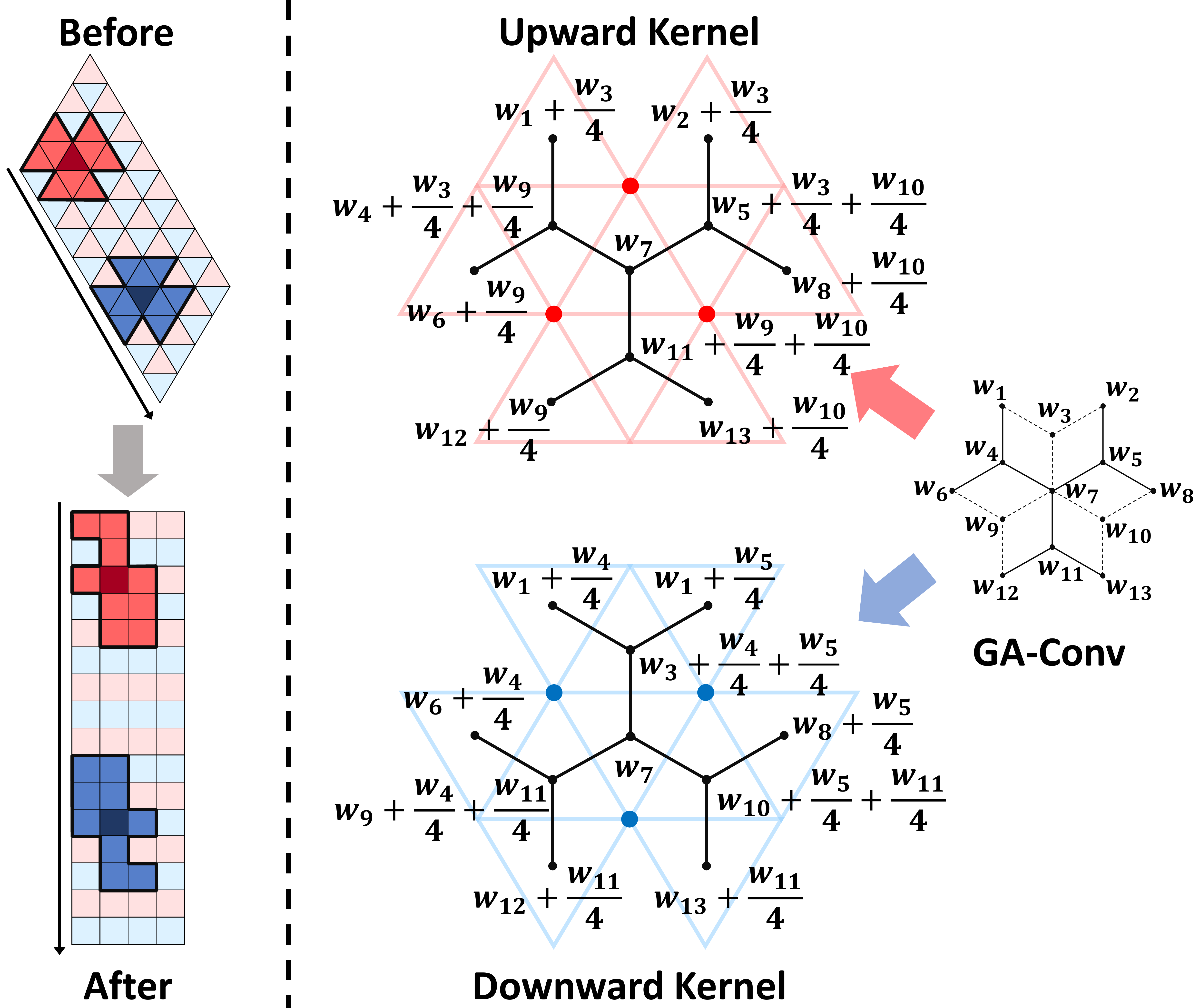}
 \vspace{-5pt}
  \caption{New kernel weight sharing. Left: proposed new data structure, Right: our kernel weight sharing scheme}
  \label{fig:ga-conv}
  \vspace{-15pt}
\end{figure}

\noindent \textbf{Data structure.} Inspired by the convolution of icosahedron data in SpherePHD, we propose  a new spherical data structure. To implement the convolution operation, SpherePHD~\cite{lee2020spherephd} uses the subdivision process of icosahedron described in Fig.~\ref{fig:icosahedron} and creates a call-table containing the indices of $N$ neighboring pixels for each pixel and then uses it to stack every neighboring pixel. After that, convolution is performed with the kernel of size $[N+1,1]$.
However, this implementation is not memory-efficient as it requires additional $N$ channels for stacking the neighboring pixels for every convolution operation.
%as they store the same pixel repetitively in call-table resulting in excessive memory usage. %for stacking neighborhood pixels
%compared with the conventional 2D convolution operation that utilizes the repetitive pattern of neighboring pixels.
% we propose a new spherical data structure that can efficiently implement convolution of meshes composing subdivision of icosahedron for 360$^\circ$ image SR. %composed subdivision of icosahedron. 
To solve this problem, we propose a new data structure that convolution operation can be directly applied without stacking the neighbors in a call-table. 
%find repetitive patterns for the efficient convolution operation. 
As shown in the left side of Fig.~\ref{fig:ga-conv}, we rearrange the original data in the direction of the arrow while transforming the triangular pixels to rectangular pixels such that conventional 2D convolution can be applied. %simply operate convolution using 2D CNN implementation.
%the convolution kernel has a repetitive convolution pattern between adjacent pixels.
Here, an upward kernel (\textcolor{red}{red kernel}) for the upward($\bigtriangleup$) aligned pixel is arranged in the odd-numbered rows, and a downward kernel (\textcolor{blue}{blue kernel}) for the downward($\bigtriangledown$) aligned pixel is arranged in the even-numbered rows. (More details are in the supplementary.)
%In such a way, it can be implemented simply through dilated convolutions.

\noindent \textbf{Kernel Weight Sharing.  } While the memory overload issue can be resolved by the proposed data structure, it is still necessary to ensure high SR performance. SpherePHD~\cite{lee2020spherephd} rotates the upward or downward kernel by 180$^{\circ}$ to obtain the same kernel shape. Therefore, it is possible to share weights of the up/downward kernels whose directions and shapes are symmetric to each other. However, as the direction of the kernel weight changes for adjacent pixels, high performance cannot be achieved if the characteristics of the texture according to the direction need to be identified.

%In the existing SpherePHD~\cite{lee2020spherephd}, a method was used to obtain the same shape by rotating the north or south kernel 180 degrees in order to share the weight of the north/south kernel whose direction and shape are symmetric to each other. 
% SpherePHD~\cite{lee2020spherephd} rotates the north or south kernel 180 degrees to obtain the same shape. As such, it is possible to share the weight of the north/south kernel whose direction and shape are symmetric to each other.
% to  method of rotating the north or south kernel 180 degrees was used to obtain the same shape. 

To solve this problem, we introduce a kernel weight sharing scheme that can geometrically align up/downward directional kernels  \emph{without rotation}.  
% weight sharing between kernels . 
%As shown in Fig.~\ref{fig:ga-conv}(b), the pixels' combination shape through which the north/south kernel passes is different, but if three vertices of the mesh located in the center are included in the pixels' combination, two shapes can be made geometrically identical. 
As shown in the right side of Fig.~\ref{fig:ga-conv}, pixel(face) combinations of two kernels, where up/downward kernels are applied, are shaped differently depending on the direction of a center pixel. 
However, if three vertices of the center pixel (denoted as \textcolor{red}{red} and \textcolor{blue}{blue} dots in the right side of Fig.~\ref{fig:ga-conv}) are included in the pixel combination as imaginary pixels, the shapes of two different up/downward pixel combinations can be made to be geometrically identical. To this end, rather than averaging and creating imaginary pixels, we distribute the kernel weight to near pixels. For the upward kernel, image pixel weight $w_3$, $w_9$, $w_{10}$ are distributed to the nearest 4 pixels except the center pixel. For downward kernel weight, $w_4$, $w_5$, $w_11$ are distributed in the same way. The detail can be seen in the right side of Fig.~\ref{fig:ga-conv}.
In this way, the feature extraction module can be applied to any pixel without rotation.

\subsection{Spherical Local Implicit Image Function (SLIIF)}

\label{image_func}
\begin{figure}[t!]
  \centering
  \includegraphics[width=.99\linewidth]{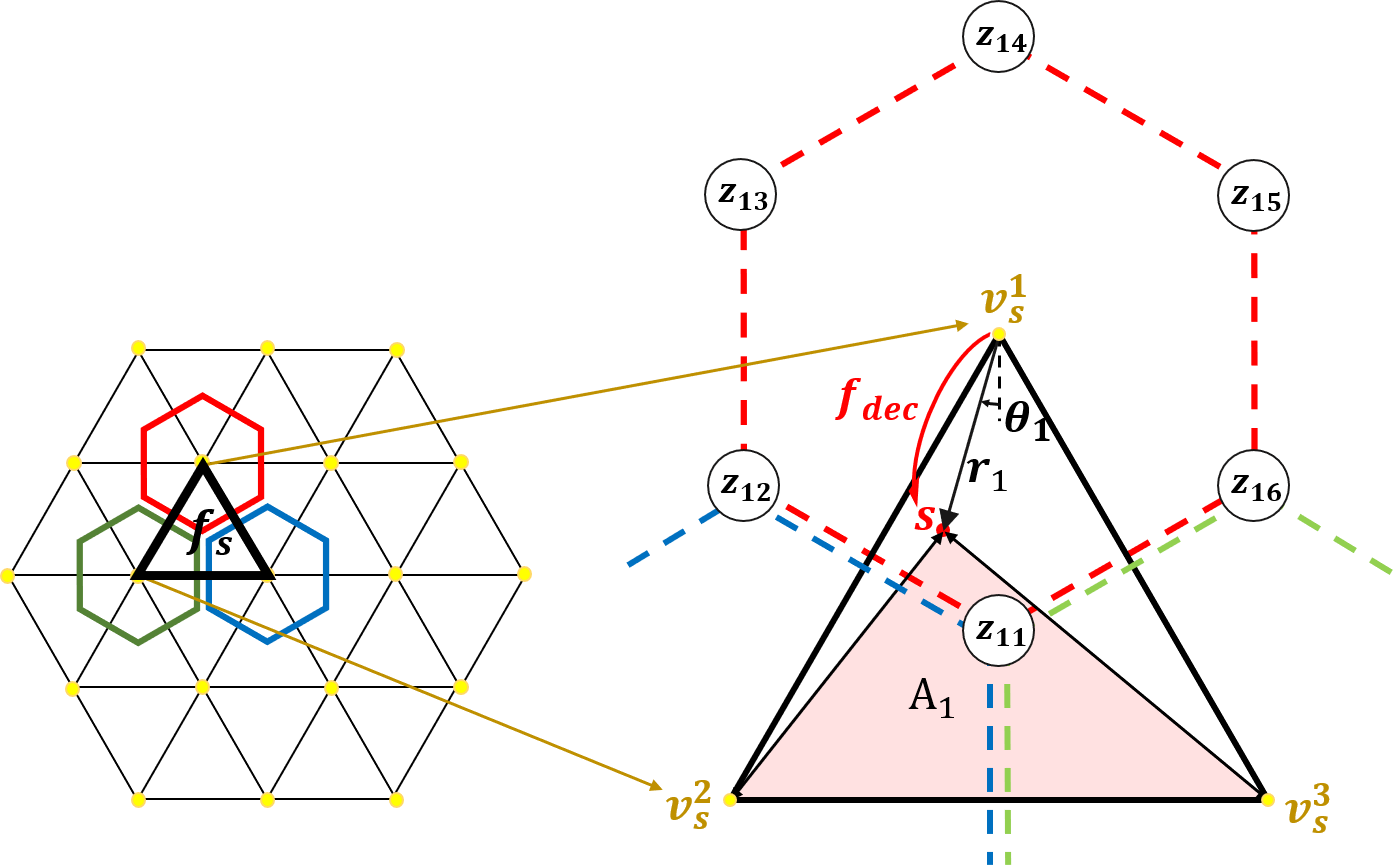}
  \vspace{-5pt}
   \caption{Spherical local implicit image function.}
   \label{fig:implict_func}
  \vspace{-10pt}
\end{figure}

\noindent \textbf{Overall Process of SLIIF. } 
With the data efficiently represented, we now describe how to super-resolve the ODIs efficiently with an arbitrary scale.
Our main idea is to predict an RGB value for an arbitrary coordinate on the unit sphere $S^2$ using the feature map extracted by using GA-Conv described in Sec.~\ref{ds_ks}.
%In this section, we present a novel method which predicts RGB value of an arbitrary point on the unit sphere $S^2$ using the feature map extracted in Sec.~\ref{ds_ks}.
Inspired by LIIF~\cite{chen2021learning}, we propose SLIIF that learns an implicit image function on $S^2$ using icosahedral faces. 
%As shown in Fig.~\ref{fig:implict_func}, SLIIF considers the characteristics of icosahedral face and ensembles 3 nearest vertex RGB values instead of ensembling 4 nearest pixel RGB values, 
SLIIF takes a spherical coordinate of the point on the unit sphere and its neighboring feature vectors as inputs and predicts the RGB value. It can be formulated as: 
% 4개의 픽셀을 레퍼런스로 잡는대신 3개의 꼭짓점을 에퍼런스로 잡았다.
% Inspired by LIIF~\cite{chen2021learning}, 우리는 icosahedral mesh상에서 implicit image function을 학습할 수 있는 SLIIF를 제안한다. 
\begin{equation}
  I(s) = f_{dec}(z, s),s\in S^2
  \label{sliif}
\end{equation}
where $f_{dec}$ is a decoding function shared with all icosahedral faces, $s$ is the point on the unit sphere $S^2$, $z$ represents a feature vector formed by concatenating neighboring feature vectors of $s$, and $I(s)$ is the predicted RGB value of $s$.

% s(or x or p_x)정의 -> s와 가장 가까운 혹은 s를 포함하는 mesh/face f_i? f_s 정의 -> f_s의 3개의 vertices v^1_s~ 정의 -> relative coordinate = (r, theta)  정의 ->  F_s1 를 v_s1를 포함하는 6개의 mesh의 feature concat으로 정의 -> 면적 A_J 정의 cell-decoding 정의 ->최종 식 정의
% x를 포함하는 mesh의 3vertice의 rgb값을 구하고 interpolation 함 (넓이 비율로) 
For the pixel in the image that can be formed by any arbitrary projection from the unit sphere, there is a corresponding point $s$ on the unit sphere $S^2$.\footnote{The coordinate of $s$ is computed by using the center point of a pixel.} The face containing $s$ is denoted as $f_s$ and three vertices surrounding $f_s$ are denoted as $v^1_s, v^2_s, v^3_s$ (see Fig.~\ref{fig:implict_func}). 
The RGB values of $s$ w.r.t. the coordinate system of three vertices are first calculated and then ensembled based on the triangular areas $A_1, A_2, A_3$ to get the final RGB value of point $s$.
The RGB value of $s$ w.r.t. each vertex $v^j_s$ is calculated with features of 6 faces containing the vertex and relative polar coordinate. 
The features of the six faces are concatenated clock-wise starting from $f_s$ to preserve geometrical consistency. Here, we denote the concatenated features as $z_j$ and the polar coordinate of $s$ with respect of $v^j_s$ as $(r_j, \theta_j)$.
To better utilize the positional information, $(r_j, \theta_j)$ are encoded with $\gamma(p) =(\sin{(2^0 \pi p)},\cos{(2^0 \pi p)},...,\sin{(2^{L-1} \pi p)},\cos{(2^{L-1} \pi p)})$ to extend the dimension of a relative coordinate introduced in ~\cite{mildenhall2020nerf,Tancik2020Fourier}.
As such, we can predict the RGB value of point $(\theta,\phi)\in S^2$, which can be formulated as:
\begin{eqnarray}
  I(\theta, \phi) = \sum_{j=1}^3 \frac{A_j}{A} \cdot f_{dec}(z_j, [\gamma (r_j),\gamma(\theta_j)]) 
\end{eqnarray}
When the pixel in the image corresponds to the vertex on $S^2$, we can still follow the aforementioned procedure because any choice of neighboring vertices results in the same RGB value due to the triangle area-based weighting. 

\label{cell_decode}

%\begin{equation}
%  RGB = \sum_{j=1}^3 \frac{A_j}{A} \cdot 
%  f_{dec}(\mathcal{F}_j,(r_j,\theta_j),(c_x,c_y))
%  \label{f2rgb}
%\end{equation}

\iffalse
\begin{table*}[t!]
\centering
\begin{tabular}{cc}
\hline
\multicolumn{1}{c}{ERP} & {Perspective} \\ \hline
  $\begin{pmatrix}
    \Delta x \hat{x} \\ \Delta y \hat{y}
  \end{pmatrix}
  =
  \begin{pmatrix}
     0 & \Delta X \sin{\theta} \\ -\Delta Y & 0
  \end{pmatrix}
  \begin{pmatrix}
    \hat{\theta} \\ \hat{\phi}
  \end{pmatrix}$
&
  $\begin{pmatrix}
    \Delta x \hat{x} \\
    \Delta y \hat{y}
  \end{pmatrix}
  =
  \begin{pmatrix}
    \alpha \Delta X \cos{\theta}\sin{\phi} & \alpha\Delta X \cos{\phi} \\
    -\alpha\Delta Y \sin{\theta} & 0
  \end{pmatrix}
  \begin{pmatrix}
    \hat{\theta} \\
    \hat{\phi}
  \end{pmatrix}$ \\\hline\hline
\multicolumn{2}{c}{Fisheye}              \\ \hline
\multicolumn{2}{c}{
  $\begin{pmatrix}
    \Delta x \hat{x} \\
    \Delta y \hat{y}
  \end{pmatrix}
  =
  \begin{pmatrix}
    \alpha \Delta X \cos{\theta}\sin{\phi} + \alpha\Delta X F_{X}(X,Y)\cos{\theta}\cos{\phi} &
    \alpha\Delta X \cos{\phi} - \alpha\Delta X F_{X}(X,Y)\sin{\phi} \\
    -\alpha\Delta Y \sin{\theta} + \alpha\Delta Y F_{Y}(X,Y)\cos{\theta}\cos{\phi} &
    - \alpha\Delta Y F_{Y}(X,Y)\sin{\phi}
  \end{pmatrix}
  \begin{pmatrix}
    \hat{\theta} \\
    \hat{\phi}
  \end{pmatrix}$

}              \\ \hline
\end{tabular}
\caption{Derivation Results of Projected vectors $\Delta x \hat{x}, \Delta y \hat{y}$ for ERP, Perspective and Fisheye Projections. }
\end{table*}
\fi

\begin{figure}[t]
  \centering
  \includegraphics[width=0.99\linewidth]{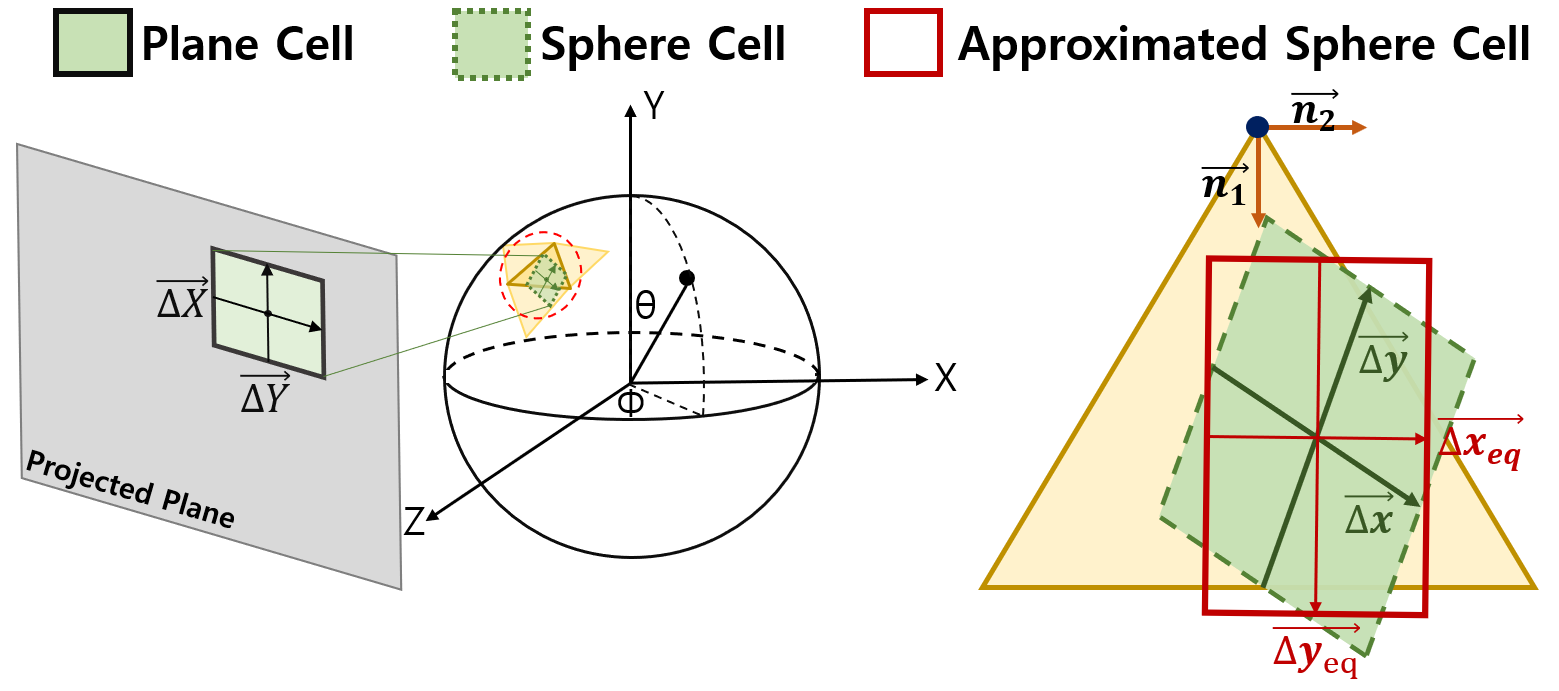}
   \vspace{-10pt}
   \caption{Sphere-oriented Cell Decoding.}
   \vspace{-10pt}
   \label{fig:celldecode}
\end{figure}

\noindent \textbf{Sphere-oriented Cell Decoding. } 
%우리는 이제 unit sphere 상의 임의의 지점에 대한 rgb value를 sLIIF를 통해 예측할 수 있고, 이를 통해 우리는 임의의 프로젝션 타입에 해당하는 unit sphere 상의 지점들을 계산하여 원하는 타입의 output을 생성할 수 있다. 하지만, 우리의 모델은 여전히 output의 단위면적에 해당하는 sphere상의 pixel area를 전혀 고려하지 않고 있다. 따라서, 우리는 임의의 프로젝션 타입으로 프로젝션된 area와 이에 해당하는 sphere상의 area사이의 상대적인 크기를 인풋으로 주어 SLIIF의 rgb value를 더 정확하게 예측할 필요가 있다. 
Through SLIIF, we can predict the RGB value for any point on $S^2$. That is, we can generate the desired HR image for any projection type by predicting the RGB value of each pixel.
%the points on the unit sphere corresponding to arbitrary projection type. 
However, SLIIF provides the RGB value only for the center of the pixel and discards the information within the pixel area except for the center value. 
 %For a 2D image, 
To handle this, LIIF~\cite{chen2021learning} defines the cell decoding value as the width and height of the query pixel of interest.
%or in this case parallelogram on mesh as the continuous representation was formed only on 2D plane. 
Nonetheless, this definition cannot be directly applied to sphere as the corresponding region on the sphere is not in a rectangular shape, and the direction of the reference vertex, where the RGB value is first calculated, continually changes. To handle this, we propose sphere-oriented cell decoding, a method considering the relation between the pixel region on the projected output and the corresponding region on $S^2$. By adding the cell decoding value as the input to SLIIF, we can fully utilize information within the pixel area.
%
%Therefore, to predict the rgb value with SLIIF more accurately, we need by giving the ratio of the area projected by any projection type and the corresponding area on the unit sphere as an input of SLIIF.
%
As shown in Fig.~\ref{fig:celldecode}, we aim at obtaining the RGB value of the rectangular pixel on the projected plane. We call this rectangular pixel a plane cell, which can be expressed using two vectors $\overrightarrow{\Delta X},\overrightarrow{\Delta Y}$. The sphere cell, the corresponding area of the plane cell on the sphere, is located in the face where the corresponding point of pixel center is located on the sphere. The sphere cell can be also expressed using two vectors $\overrightarrow{\Delta x},\overrightarrow{\Delta y}$. The relation between $\overrightarrow{\Delta X},\overrightarrow{\Delta Y}$ and $\overrightarrow{\Delta x},\overrightarrow{\Delta y}$ depends on the projection type and location of the pixel center (refer to the supplementary for details). 

For the geometrical consistency of the orders of the concatenated features, relative coordinate of $s$, and cell decoding value among pixels, we need to define new axis vectors, $\overrightarrow{n_1}$ and $\overrightarrow{n_2}$, invariant to face orientation.
%
%we define two new axis vectors $\overrightarrow{n_1}$ and $\overrightarrow{n_2}$. 
The unit vector $\overrightarrow{n_1}$ is defined as a vector between the reference vertex and the face center, and the unit vector $\overrightarrow{n_2}$ is defined by 90 degree counter-clockwise rotation of $\overrightarrow{n_1}$. To get the height and width based on this coordinate system, we approximate the parallelogram sphere cell to the axis-aligned rectangle. The approximated sphere cell is defined as a rectangle which can be expressed using two vectors  $\overrightarrow{\Delta x_{eq}}, \overrightarrow{\Delta y_{eq}}$ that has the same area as the parallelogram sphere cell and the largest intersection area with the parallelogram sphere cell. %$A=\underset{X\in\{X \: \boldsymbol{|}\: |X|=|B|\}}{argmax}|X\cap B|$
%the rectangular pixel on the projected plane can be expressed using two vectors $(\Delta X \hat{X},\Delta Y \hat{Y})$. On the mesh, where corresponding point of pixel center is located, the corresponding parallelogram area of the rectangle pixel can also be expressed using two vectors $(\Delta x \hat{x},\Delta y \hat{y})$. The relation between $(\Delta X \hat{X},\Delta Y \hat{Y})$ and $(\Delta x \hat{x},\Delta y \hat{y})$ depends on the projection type and the location of the pixel center.
% w.r.t. the vertex aiming to predict RGB value, 
Based on the approximated rectangular sphere cell, we finally formulate the sphere-oriented cell decoding value as:

\begin{equation}
  \begin{matrix}
    \begin{pmatrix}
        \overrightarrow{\Delta x} \\ \overrightarrow{\Delta y}
    \end{pmatrix}
    \approx
    \begin{pmatrix}
        \overrightarrow{\Delta x_{eq}} \\ \overrightarrow{\Delta y_{eq}}
    \end{pmatrix}    
    =
    \begin{pmatrix}
        c_x\overrightarrow{n_{1}} \\c_y\overrightarrow{n_{2}}
    \end{pmatrix}     
    
    % \Delta x \hat{x} \approx \Delta x_{eq} \hat{x}_{eq} = \alpha_1'\overrightarrow{n_{1}} \\
    % \Delta y \hat{y} \approx \Delta y_{eq} \hat{y}_{eq} = \beta_2'\overrightarrow{n_{2}}
  \end{matrix}
\end{equation}

\begin{equation}
    \Rightarrow c = [c_{x}, c_{y}]
    = \left(\frac{|\overrightarrow{\Delta x_{eq}}|} {|\overrightarrow{n_{1}}|},\frac{|\overrightarrow{\Delta y_{eq}}|} {|\overrightarrow{n_{2}}|}\right)
\end{equation}

As a result, we can predict the RGB value $I(X,Y)$ for any point on the projected plane based on the following equation,
\begin{equation}
    \begin{aligned}
        I(X,Y) &= I(\theta,\phi,c)  \\
        &=\sum_{j=1}^3 \frac{A_j}{A} \cdot f_{dec}(z_j, [\gamma (r_j),\gamma(\theta_j)],[c_x,c_y]) 
    \end{aligned}
\end{equation}

\subsection{Loss Function}

\begin{figure}
\includegraphics[width=1.0\linewidth]{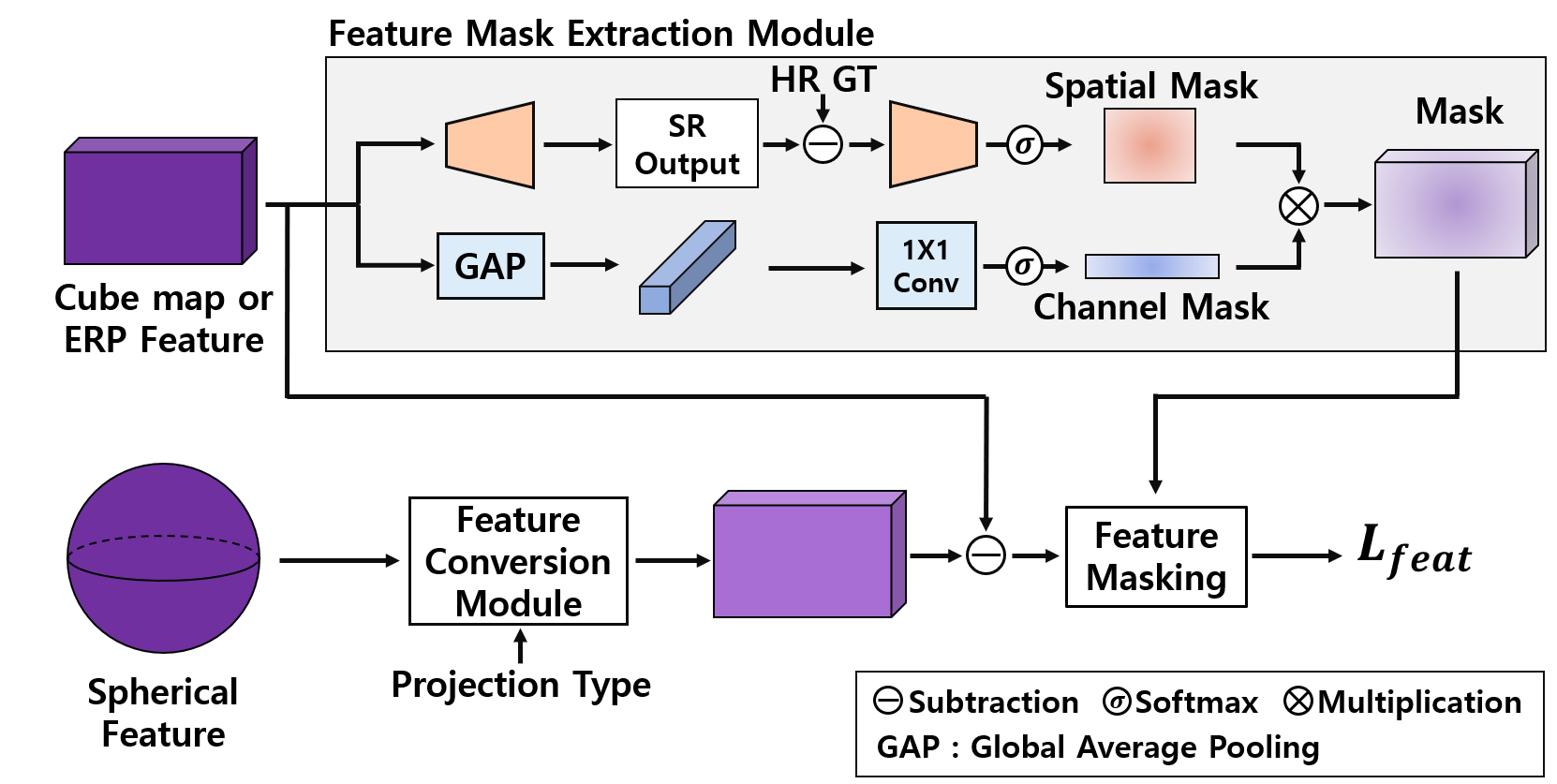}
\caption{The proposed feature loss between the spherical and ERP features.}
\label{fig:loss}
 \vspace{-5pt}
\end{figure}

\begin{table*}[t!]
\small
\renewcommand{\arraystretch}{0.85}
\renewcommand{\tabcolsep}{6pt}
\centering
\caption{ERP SR results on the ODI-SR and SUN 360 Panorama Dataset. \textbf{Bold} indicates the best results.}
\vspace{-4pt}
\begin{tabular}{c||cccc||cccc}
\hline 
% \thickhline
Scale       & \multicolumn{4}{c||}{x8}                                                                     & \multicolumn{4}{c}{x16}                                                                             \\ \hline
\multirow{2}{*}{Method} & \multicolumn{2}{c|}{ODI-SR} & \multicolumn{2}{c||}{SUN 360 Panorama}  & \multicolumn{2}{c|}{ODI-SR}  & \multicolumn{2}{c}{SUN 360 Panorama}  \\ \cline{2-9} 
                        & \multicolumn{1}{c|}{WS-PSNR} & \multicolumn{1}{c|}{WS-SSIM} & \multicolumn{1}{c|}{WS-PSNR} & WS-SSIM & \multicolumn{1}{c|}{WS-PSNR} & \multicolumn{1}{c|}{WS-SSIM} & \multicolumn{1}{c|}{WS-PSNR} & WS-SSIM \\ \hline\hline
Bicubic                 & \multicolumn{1}{c|}{19.64}   & \multicolumn{1}{c|}{0.5908}  & \multicolumn{1}{c|}{19.72}   & 0.5403  & \multicolumn{1}{c|}{17.12}   & \multicolumn{1}{c|}{0.4332}  & \multicolumn{1}{c|}{17.56}   & 0.4638  \\ \hline
SRCNN~\cite{Dong2016Image}                 & \multicolumn{1}{c|}{20.08}   & \multicolumn{1}{c|}{0.6112}  & \multicolumn{1}{c|}{19.46}   & 0.5701  & \multicolumn{1}{c|}{18.08}   & \multicolumn{1}{c|}{0.4501}  & \multicolumn{1}{c|}{17.95}   & 0.4684  \\ \hline
VDSR~\cite{kim2016accurate}                    & \multicolumn{1}{c|}{20.61}   & \multicolumn{1}{c|}{0.6195}  & \multicolumn{1}{c|}{19.93}   & 0.5953  & \multicolumn{1}{c|}{18.24}   & \multicolumn{1}{c|}{0.4996}  & \multicolumn{1}{c|}{18.21}   & 0.4867  \\ \hline
LapSRN~\cite{lai2017deep}                  & \multicolumn{1}{c|}{20.72}   & \multicolumn{1}{c|}{0.6214}  & \multicolumn{1}{c|}{20.05}   & 0.5998  & \multicolumn{1}{c|}{18.45}   & \multicolumn{1}{c|}{0.5161}  & \multicolumn{1}{c|}{18.46}   & 0.5068  \\ \hline
MemNet~\cite{tai2017Memnet}                  & \multicolumn{1}{c|}{21.73}   & \multicolumn{1}{c|}{0.6284}  & \multicolumn{1}{c|}{21.08}   & 0.6015  & \multicolumn{1}{c|}{20.03}   & \multicolumn{1}{c|}{0.5411}  & \multicolumn{1}{c|}{19.88}   & 0.5401  \\ \hline
MSRN~\cite{li2018multiscale}                    & \multicolumn{1}{c|}{22.29}   & \multicolumn{1}{c|}{0.6315}  & \multicolumn{1}{c|}{21.34}   & 0.6002  & \multicolumn{1}{c|}{20.05}   & \multicolumn{1}{c|}{0.5416}  & \multicolumn{1}{c|}{19.87}   & 0.5316  \\ \hline
EDSR~\cite{lim2017enhanced}                    & \multicolumn{1}{c|}{23.97}   & \multicolumn{1}{c|}{0.6417}  & \multicolumn{1}{c|}{22.46}   & 0.6341  & \multicolumn{1}{c|}{21.12}   & \multicolumn{1}{c|}{0.5698}  & \multicolumn{1}{c|}{21.06}   & 0.5645  \\ \hline
D-DBPN~\cite{haris2018ddbpn}                  & \multicolumn{1}{c|}{24.15}   & \multicolumn{1}{c|}{0.6573}  & \multicolumn{1}{c|}{23.70}   & 0.6421  & \multicolumn{1}{c|}{21.25}   & \multicolumn{1}{c|}{0.5714}  & \multicolumn{1}{c|}{21.08}   & 0.5646  \\ \hline
RCAN~\cite{zhang2018image}                    & \multicolumn{1}{c|}{24.26}   & \multicolumn{1}{c|}{0.6628}  & \multicolumn{1}{c|}{23.88}   & 0.6542  & \multicolumn{1}{c|}{21.94}   & \multicolumn{1}{c|}{0.5824}  & \multicolumn{1}{c|}{21.74}   & 0.5742  \\ \hline
EBRN~\cite{qiu2019Embedded}                    & \multicolumn{1}{c|}{24.29}   & \multicolumn{1}{c|}{0.6656}  & \multicolumn{1}{c|}{23.89}   & 0.6598  & \multicolumn{1}{c|}{21.86}   & \multicolumn{1}{c|}{0.5809}  & \multicolumn{1}{c|}{21.78}   & 0.5794  \\ \hline
360-SS~\cite{ozcinar2019super}                  & \multicolumn{1}{c|}{21.65}   & \multicolumn{1}{c|}{0.6417}  & \multicolumn{1}{c|}{21.48}   & 0.6352  & \multicolumn{1}{c|}{19.65}   & \multicolumn{1}{c|}{0.5431}  & \multicolumn{1}{c|}{19.62}   & 0.5308  \\ \hline
LAU-Net~\cite{deng2021lau}                 & \multicolumn{1}{c|}{24.36}   & \multicolumn{1}{c|}{\textbf{0.6801}}  & \multicolumn{1}{c|}{24.02}   & 0.6708  & \multicolumn{1}{c|}{22.07}   & \multicolumn{1}{c|}{0.5901}  & \multicolumn{1}{c|}{21.82}   & 0.5824  \\ \hline
SphereSR(Ours)          & \multicolumn{1}{c|}{\textbf{24.37}}   & \multicolumn{1}{c|}{0.6777}  & \multicolumn{1}{c|}{\textbf{24.17}}   & \textbf{0.6820}  & \multicolumn{1}{c|}{\textbf{22.51}}   & \multicolumn{1}{c|}{\textbf{0.6370}}  & \multicolumn{1}{c|}{\textbf{21.95}}   & \textbf{0.6342}  \\ 
\hline
% \thickhline
\end{tabular}
\label{result_table}

\end{table*}

\label{loss_func}

We train the proposed framework using two loss terms. First, we use the multi-scale L1 loss. With the L1 loss defined at the multiple scales, our framework can learn more about various relative coordinates and cell decoding values.
%
% so the current task of extracting results for arbitrary projection type at the time of testing must be required.
%First, we propose a feature similarity loss to give a similarity between features extracted from input of other projection types(ERP and cube map) and features extracted from sphere. As shown in fig.~\ref{fig:loss}, the spherical feature go through a feature conversion module with the structure of the SLIIF, which allows conversion to features similar to those of other projection types. In addition, it is designed so that the gradient backward can be achieved only for important areas that can actually help by creating a learnable feature mask considering the characteristics of the channel and spatial of the feature.
Second, we design a feature loss module to measure the similarity between the features extracted from the sphere and other projection types. 
% to induce the network to learn good features from pre-trained SR network of other projection type and extract the accurate RGB values. % for arbitrary projection type. 

As shown in fig.~\ref{fig:loss}, we design a feature mask from the ERP or cube map feature. The spatial part of the mask is generated from the difference between the predicted SR ODI and HR ground truth. The channel part of the mask is generated from the features via channel-wise global average pooling. In this way, we get the feature mask emphasizing the relevant part with high accuracy. 
Also, the spherical features are converted to the shapes of other projection types via the SLIIF feature conversion module. Finally, the converted features are then subtracted and masked to formulate the feature loss $L_{feat}$. The total loss is as follows.

% the mask is designed to focus on the strength of other projection type. from the difference of super resolved image and ground truth, spatial mask is generated. using GAP, channel wise mask is generated. multiplication of channel and spatial mask results to the final mask form. this mask aim to activate strong part of feature where sr and gt is similar and deactivate poor part. 좋은 부분/.activete된부분에대해서만 .sphere feature와 pther projection types' feture 의 차이를 최소화해 좋은부부능ㄹ 학습한다. 
%we aim to leran from features where it have high perfromance. error map을 loss로 주는게 이상적이다.
%Softmax operation is included to prevent mask conversing to zero 

%go through a feature conversion module with the structure of the mesh-based LIIF, which allows switching to features similar to those of other projection types. In addition, it is designed so that the gradient backward can be achieved only for important areas that can actually help by creating a learnable feature mask considering the characteristics of the channel and spatial of the feature.

%multiscale and ERP loss is enough to gain ability to produce arbitary projection result

\begin{equation}
    Loss = \frac{1}{N} \sum_{j=1}^N \parallel I_j^{est} - I_j^{gt}\parallel_1  + \lambda L_{feat}
\end{equation}
%%%%%%%%%%%%%%%%%% Experiment %%%%%%%%%%%%%%%%%%
%ODI-dataset training/test result
%arbitrary projection test result
%ablation studies

\section{Experiments}

\subsection{Dataset and Implementation}
%우리는 ODI-SR dataset[lau-net]을 이용하여 네트워크를 training하고 test한다. Training을 위해, ODI-SR training dataset 800장 중 750장을 사용하고, validation을 위해 나머지 50장을 사용한다. Test를 위해, 우리는 ODI-SR test dataset 100장과 SUN360 Panorama Database[] 100장을 사용한다. 
%ODI dataset HR 이미지의 resolution은 1024 by 2048이며, scale factor x8과 x16에 대한 super-resolution수행을 위해 training을 진행한다. 이때, feature loss를 위한 ERP, fisheye super-resolution network는 EDSR 구조를 가지며, ODI dataset에서 미리 pretrain한 후 freeze한다. 우리 모델,SphereSR,은 fig.[2]에서 묘사된 icosahedral meshes를 input으로 가지며, x8 및 x16 scale에 대한 input level은 각각 5와 6으로 설정하였다. 또한, 우리 모델의 feature extractor는 GA-conv로 구현한 Spherical CNN model이며, EDSR과 동일한 구조를 가지고 있고, feature dimension은 128이다.  RGB prediction을 위한 decoding function f_dec는 5-layer MLP이며, hidden dimension은 256이다. 우리는 learning rate 0.0001인 Adam Optimizer[]를 이용하여batch size 1에서 500 epochs만큼 train하고, 400 epoch에서 learning rate를 10배 감소시킨다. 또한 train시에, 처음에는 feature loss scale parameter \lambda를 0으로 두어 먼저 학습시키고, 100epoch 이후 \lambda를 0.3으로 주고 재학습시킨다. 

We train and test SphereSR using the ODI-SR dataset~\cite{deng2021lau} and SUN360 panorama dataset~\cite{xiao2012recognizing}. For training, 750 out of 800 ODI-SR training images are used and the remaining 50 images are used for validation. For testing, we use 100 images from the ODI-SR test dataset and another 100 images from the SUN360 panorama dataset.
The resolution of an HR ODI is 1024$\times$2048, and training is performed for the scale factor $\times$8 and $\times$16. 
As shown in Fig.~\ref{fig:icosahedron}, SphereSR takes an image on icosahedron as an input converted from LR ODIs, and the icosahedron subdivision levels for the scale $\times$8 and $\times$16 are set to 5 and 6, respectively. The spherical feature extractor is implemented using GA-conv based on EDSR~\cite{lim2017enhanced} and outputs a feature with 128 dimensions. The RGB prediction decoding function $f_{dec}$ is a 5-layer MLP, and the hidden layer dimension is 256. The EDSR model pre-trained with the ODI-SR dataset is used to extract features from ERP and cube map images for feature loss. 
%training과정에서, 우리의 모델은 세 가지의 서로다른 projection type을 input으로 사용하므로, LR ODI를 nearest interpolation을 사용하여 해당 input type으로 변형해준다.
%We use EDSR~\cite{lim2017enhanced} as the baseline networks for the ERP and fisheye image SR. The baseline network is pre-trained in advance in the ODI dataset and then is fixed. 
SphereSR is trained for 500 epochs using the Adam Optimizer with batch size 1. 
The initial learning rate is set to 0.0001 and reduced by 10 times after 400 epochs.
The feature loss scale parameter $\lambda$ is set to 0 for the first 100 epochs and set to 0.3 for the rest of the training.

%% 설명 순서가 맞는지 모르겠음

\subsection{Evaluation on ERP}

%우리는 LAU-Net 및 여러 SR model들과의 비교를 위해, ODI dataset에서 scale 8,16배에 대한 ERP super-resolution를 진행하였고, test는 ODI-SR dataset 그리고 SUN360 Panorama dataset에서 진행했다. 우리는 ODI SISR을 위한 네트워크인 360-SS[], LAU-Net[]과 perspective image에 대한 SISR을 위한 9가지 모델[]과 우리의 모델 SphereSR과 결과를 비교했다. 우리는 metric으로 ODI를 위한 수치인 WS-PSNR[]과 WS-SSIM[]을 이용하였다.
%Table 1에서 볼 수 있듯이, 우리 모델, SphereSR은 ODI-SR과 SUN360 Panorama dataset에 대해서 x8, x16 SR에 대한 대부분의 성능이 다른 모델보다 높은 것을 알 수 있다. 

% For comparison with LAU-Net and several SR models, we conduct ERP SR for the 
% scale of $\times$8 and $\times$16 on the ODI data, and 
We use the ODI-SR and SUN360 Panorama datasets for evaluation. We compare SphereSR with 9 models for 2D SISR, including SRCNN~\cite{Dong2016Image}, VDSR~\cite{kim2016accurate}, LapSRN~\cite{lai2017deep}, MemNet~\cite{tai2017Memnet}, MSRN~\cite{li2018multiscale}, EDSR~\cite{lim2017enhanced}, D-DBPN~\cite{haris2018ddbpn}, RCAN~\cite{zhang2018image}, EBRN~\cite{qiu2019Embedded} and 2 models for ODI-SR, \ i.e., 360-SS~\cite{ozcinar2019super} and LAU-Net~\cite{deng2021lau}.   
% perspective image, which are networks for ODISR, and our model SphereSR. 
We use WS-PSNR~\cite{Zhou2018WeightedtoSphericallyUniformSO} and WS-SSIM~\cite{Zhou2018WeightedtoSphericallyUniformSO} as the evaluation metrics. 
% As shown in Table 1 for $\times$8 and $\times$16 SR, our model, SphereSR, has higher performance for most of x8 and x16 SRs for ODI-SR and SUN360 Panorama Dataset.
%~\ref{table:result_table}에서 볼 수 있듯이, x8 upscaling에 대한 결과를 보면 ODI-SR의 WS-PSNR 그리고 SUN360 Panorama의 WS-PSNR/WS-SSIM에서 SphereSR이 가장 높은 성능을 가진다. 또한, ODI-SR의 WS-SSIM은 LAU-Net 이후로 두 번째로 가장 높은 성능을 가진다. 두 번째로 x16 upscaling에 대한 결과를 보면, ODI-SR 및 SUN360 Panorama 두 가지의 dataset에 대한 WS-PSNR과 WS-SSIM 성능에서 다른 method와 비교했을 때, SphereSR이 가장 높은 성능을 가진다. 

\noindent \textbf{Quantitative results.} Table~1 shows the quantitative comparison for $\times$8 and $\times$ 16 SR on the ODI-SR and SUN 360 panorama datasets. As can be seen, SphereSR outperforms all the compared methods on both datasets except the case of $\times$8 SR on the ODI-SR dataset, where SphereSR shows comparable performance with LAU-Net. 
However, for $\times$16 SR, SphereSR has the higher performance compared to LAU-Net in WS-PSNR and WS-SSIM performance on both ODI-SR and SUN360 panorama datasets. 

\begin{figure*}[]
  \centering
  \includegraphics[width=0.99\linewidth]{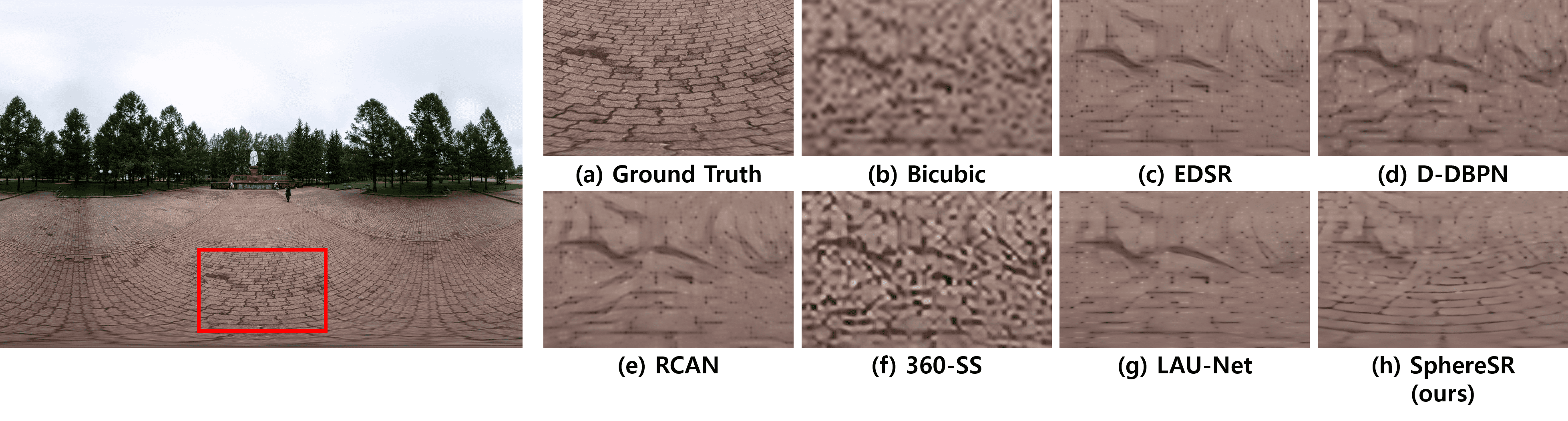}
  \vspace{-16pt}
   \caption{Visual comparisons of x8 SR results of different methods on ODI-SR dataset.}
   \label{fig:ERP_result}
 \vspace{-10pt}
\end{figure*}

\noindent \textbf{Qualitative comparison.} Figure~\ref{fig:ERP_result} shows the visual comparison of the $\times$8 SR images on the ODI-SR dataset. As can be seen, SphereSR reconstructs clear textures and more accurate structures, while other compared methods suffer from the problems of blurred edges or distorted structures.
From the visual comparison, we can conclude that SphereSR produces the texture of repeated patterns more accurately than ERP networks.

%\begin{figure*}[]
%  \centering
%  \includegraphics[width=1.0\linewidth]{figure/result_TBD.png}
%   \caption{Visual comparisons of x8 SR results of different methods on ODI-SR dataset.}
%   \label{fig:onecol}
%\end{figure*}

%\input{tables/ERP_result_img}

%우리는 ERP image로 구성된 ODI-SR dataset에서 학습한 모델, SphereSR이 임의의 projection type에 대해서도 좋은 성능을 가질 수 있는지 확인하기 위해 두 가지 실험을 진행한다. 우리는 첫 번째로 512 by 512 크기의 FOV 90$^{\circ}$ perspective image로의 변환에 대한 실험을 진행하고, 우리는 두 번째로 1024 by 1024 크기의 FOV 180$^{\circ}$ fisheye image로의 변환에 대한 실험을 진행한다. 또한 우리는 fisheye projection의 여러 종류 중 하나인 circular fisheye projection을 사용한다. 
%다른 모델들과의 성능비교를 위해서, output으로 나온 erp image를 cv2 함수를 이용하여 원하는 projection type에 대한 bicubic interpolation을 수행한다. 또한, 성능 수치 비교를 위해 ERP GT image도 원하는 projection type으로 bicubic interpolation한다. 우리는 수치비교를 위한 metric으로 PSNR과 SSIM을 이용했고, 전방향의 성능을 고려하기 위해 임의의 5개 방향을 선정하고, 해당하는 방향에 맞는 projection output을 생성한 후 metric 수치의 평균값을 계산하였다.

%또한, SLIIF의 효과를 입증하기 위해, 우리는 기존 SphereSR에서 SLIIF를 제거하고, PixelShuffle module을 추가하여 SR을 수행한 뒤, 원하는 projection type으로 bicubic interpolation을 진행하고 성능을 비교해보았다. 
\subsection{SR for Other  Projection Types}
In this section, we verify whether the proposed SphereSR, trained using the ERP images on the ODI-SR dataset, can perform well for any projection type. We first conduct an experiment on the conversion to FOV 90$^{\circ}$ perspective image with a size 512$\times$512. We then conduct another experiment on the conversion to FOV 180$^{\circ}$ fisheye image with a size 1024$\times$1024. In addition, we use circular fisheye projection, one of several types of fisheye projections.
To compare with other SR models, we use bicubic interpolation to convert to the desired projection type. The ERP GT image is also interpolated using the bicubic method to the desired projection type for performance evaluation. We use PSNR and SSIM as the evaluation metrics. Note that we select five random directions, generate a projection output suitable for the corresponding direction, and calculate the mean value for PSNR and SSIM. 

% In addition, to prove the effect of SLIIF, we removed SLIIF from the existing SphereSR, added a PixelShuffle module to perform SR, and then performed bicubic interpolation with the desired projection type and compared the performance.

\begin{figure}[t!]
  \centering
  \includegraphics[width=0.99\linewidth]{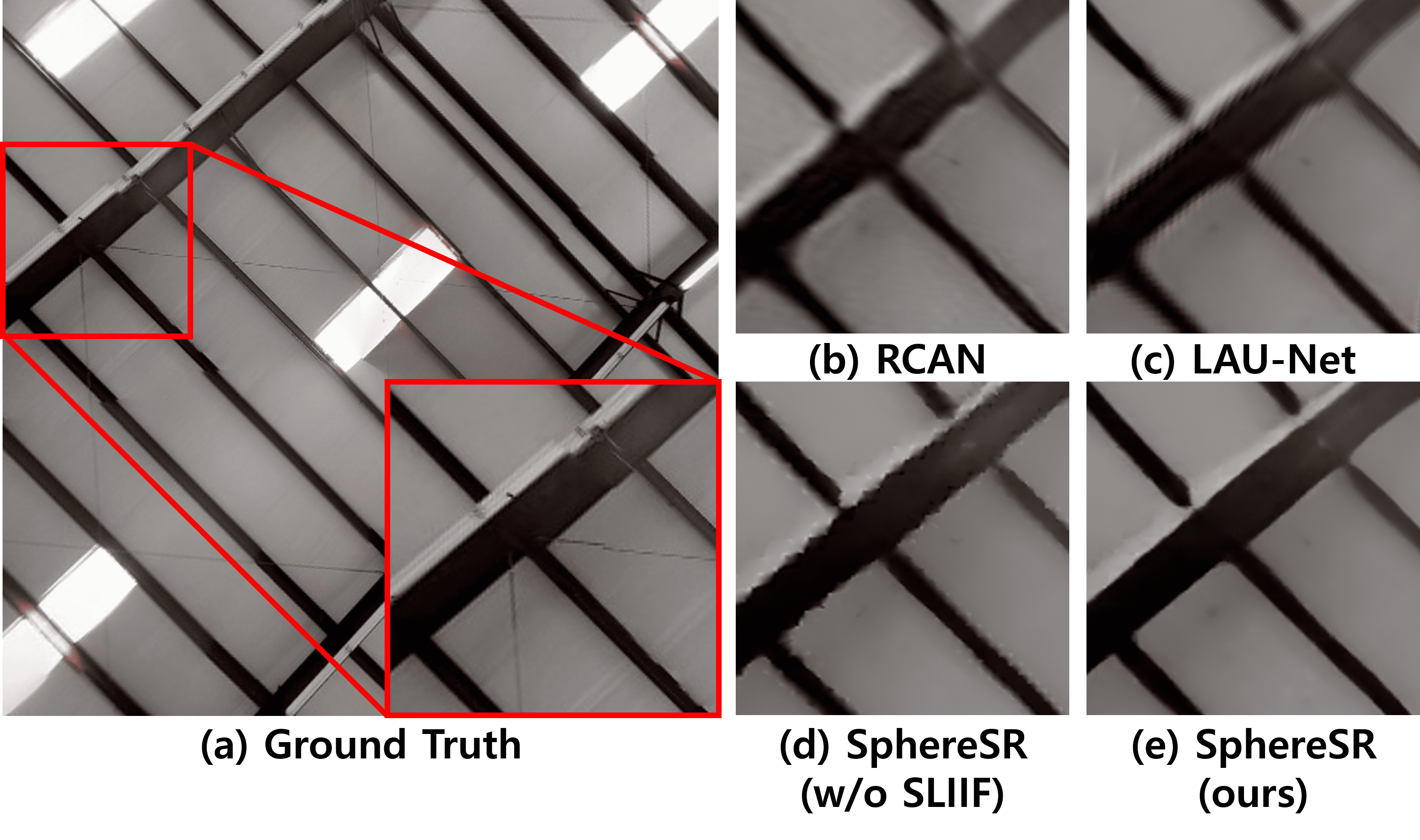}
   \vspace{-7pt}
   \caption{Visual comparison for x8 SR of perspective images on ODI-SR dataset.}
   \label{fig:Perspective_result}
   \vspace{-5pt}
\end{figure}

\begin{figure}[t!]
  \centering
  \includegraphics[width=0.99\linewidth]{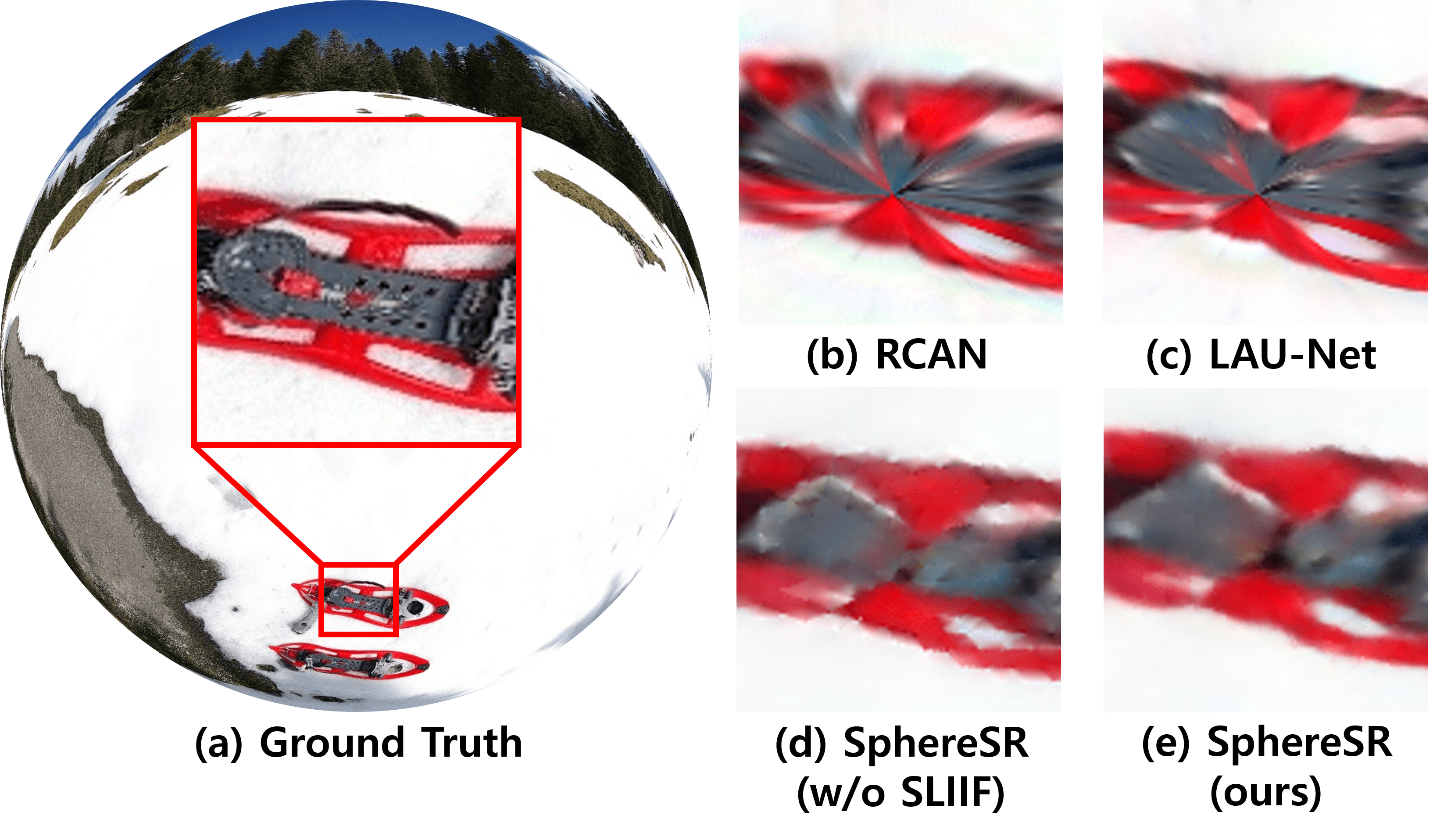}
    \vspace{-7pt}
   \caption{Visual comparison for x8 SR of fisheye images on ODI-SR dataset}
   \label{fig:Fisheye_result}
   \vspace{-12pt}
\end{figure}

\noindent \textbf{Perspective Image. } 
%As shown in the projection results of Table.~\ref{arbitrary} to the perspective image, it can be seen that SphereSR has the highest performance in both the ODI-SR dataset and the SUN360 Panorama dataset. In addition, it can be seen that the performance of SphereSR without SLIIF is not high.
%Table.~\ref{table:arbitrary}의 perspective image로의 projection결과를 보면, ODI-SR dataset과 SUN 360 Panorama dataset 결과에서 모두 SphereSR이 가장 높은 성능을 가지는 것을 확인할 수 있다. 또한, SLIIF를 사용하지 않은 SphereSR의 경우 성능이 높지 않음을 알 수 있다.
% Table.~\ref{arbitrary}의 perspective projection에 대한 결과를 보면, ODI-SR과 SUN360 Dataset에서 WS-PSNR/WS-SSIM 성능 모두 SphereSR이 가장 높은 수치를 가진다. 이전의 ODI를 위한 SR network 중 가장 높은 성능을 가진 네트워크인 LAU-Net~\cite{deng2021lau}은 ODI-SR dataset에 대한 PSNR 성능이 26.39이다. 그러나 SphereSR은 ODI-SR dataset에 대한 PSNR 성능이 26.76으로 가장 높은 성능을 가진다. 이와 더불어, SphereSR without SLIIF와 비교했을 때, SphereSR with SLIIF의 PSNR 수치가 ODI-SR은 0.1, SUN360 Panorama는 0.14 만큼 더 더 높게 측정된다.
Table.~\ref{arbitrary} shows the quantitative results for perspective image SR. SphereSR again achieves the best performance on both ODI-SR and SUN360 datasets. LAU-Net~\cite{deng2021lau} achieves the PSNR values of 26.39dB and 24.33dB on both datasets, respectively. By contrast, our method significantly surpasses LAU-Net and achieves the highest PSNR values of 26.76dB and 24.46dB on both datasets, respectively. 
In addition, when removing the SLIIF component in SphereSR, the PSNR values drop by 0.1dB and 0.14dB, respectively.
In Fig.~\ref{fig:Perspective_result}, we show a visual comparison between SphereSR with SLIIF, SphereSR without SLIIF, RCAN~\cite{zhang2018image}  and LAU-Net~\cite{deng2021lau}. 
As can be seen, SphereSR reconstructs clear straight lines and textures rather than other compared methods RCAN(b) and LAU-Net(c). Also, as a comparison for the usage of SLIIF(d,e), triangle-shaped artifacts are created when SLIIF is not used(d), but it can be confirmed that clear straight lines are created when SLIIF is used(e).

%In Fig.~\ref{fig:Perspective_result}, we show a visual comparison between SphereSR with SLIIF, SphereSR without SLIIF, RCAN~\cite{zhang2018image}  and LAU-Net~\cite{deng2021lau}. Specifically, we cropped the area to see the SR results at the south pole.
%As can be seen, RCAN(b) and LAU-Net(c) generates an inappropriate texture with several lines rushing to the south pole. On the other hand, SphereSR(w/o SLIIF)(d) and SphereSR(w/ SLIIF)(e) do not create such a texture, and in the case of (e), it eliminates the triangle-shaped artifact generated in (d).

\noindent \textbf{Fisheye. } 
%As shown in the projection result of Table.~\ref{table:arbitrary} to the fisheye image, we can see that it has a high value in all performance evaluation values, just like the perspective image. Also, it can be seen that the performance of SphereSR without SLIIF is not high.
%To this end, We can conclude that the role of SLIIF is important when performing SR for arbitrary projection type.
%Table.~\ref{table:arbitrary}의 fisheye image로의 projection결과를 보면, perspective image와 마찬가지로 모든 성능 평가 수치에서 갖아 높은 수치를 가지는 것을 확인할 수 있다. 또한, SLIIF를 사용하지 않은 SphereSR의 경우 성능이 높지 않다는 것 또한 알 수 있다.
%우리는 이로서 arbitrary projection type에 대한 SR을 수행할 때, SLIIF의 역할이 중요하다는 것을 알 수 있다.
% Table.~\ref{table:arbitrary}의 Fisheye projection 결과를 살펴보면, perspective projection결과와 마찬가지로 ODI-SR dataset 및 SUN360 Panorama dataset에 대한 PsNR/SSIM 수치에서 SpehreSR이 가장 높은 성능을 가지는 것을 확인할 수 있다. 이전의 SISR m
% ethod 중 ODI-SR dataset의 PSNR 성능이 가장 높은 model은 RCAN이고 PSNR 수치는 24.40이다. 또한 이전 모델 중 SUN360 Panorama dataset에서 PSNR이 가장 높은 model은 LAU-Net이고 PSNR 수치는 24.97이다. 
% 이와 더불어, SphereSR without SLIIF와 비교했을 때, SphereSR with SLIIF의 PSNR 수치가 ODI-SR 및 SUN360 Dataset에서 모두 0.14 올라간다.
Table.~\ref{arbitrary} shows the quantitative results for fisheye image SR. It can be seen that SphereSR has the highest performance regarding the PSNR and SSIM values on the ODI-SR and SUN360 panorama datasets.
% As in the results of the permanent projection. 
Among the methods for 2D SISR, RCAN achieves the second-highest PSNR value of 24.40dB on the ODI-SR dataset. On the SUN360 panorama dataset, LAU-Net achieves the second-highest PSNR of 24.97dB. Our method shows the highest PSNR and SSIM values, showing the best SR performance.
In Fig.~\ref{fig:Fisheye_result}, we show a visual comparison between SphereSR with SLIIF, SphereSR without SLIIF, RCAN~\cite{zhang2018image} and LAU-Net~\cite{deng2021lau}. Specifically, we crop the area to view the SR results at the south pole.
As can be seen, RCAN(b) and LAU-Net(c) generate inappropriate textures with several lines rushing to the south pole. On the other hand, SphereSR(w/o SLIIF)(d) and SphereSR(w/ SLIIF)(e) do not have such a problem. Moreover, In the case of (e), it eliminates the triangle-shaped artifact generated in (d).

\begin{table*}[t!]
\small
\renewcommand{\arraystretch}{0.9}
\renewcommand{\tabcolsep}{9pt}
\centering
\caption{Perspective and fisheye SR results on the ODI-SR and SUN 360 Panorama Dataset. \textbf{Bold} indicates the best results. 
% Results are evaluated by WS-PSNR and WS-SSIM.
} 
\vspace{-3pt}
\begin{tabular}{c||cccc||cccc}
\hline 
% \thickhline
Projection Type       & \multicolumn{4}{c||}{Perspective}                                                                     & \multicolumn{4}{c}{Fisheye}                                                                             \\ \hline
FOV      & \multicolumn{4}{c||}{90}                                                                     & \multicolumn{4}{c}{180}                                                                             \\ \hline
\multirow{2}{*}{Method} & \multicolumn{2}{c|}{ODI-SR} & \multicolumn{2}{c||}{SUN 360 Panorama}  & \multicolumn{2}{c|}{ODI-SR}  & \multicolumn{2}{c}{SUN 360 Panorama}  \\ \cline{2-9} 
                        & \multicolumn{1}{c|}{PSNR} & \multicolumn{1}{c|}{SSIM} & \multicolumn{1}{c|}{PSNR} & SSIM & \multicolumn{1}{c|}{PSNR} & \multicolumn{1}{c|}{SSIM} & \multicolumn{1}{c|}{PSNR} & SSIM \\ \hline\hline
Bicubic             & \multicolumn{1}{c|}{25.40}   & \multicolumn{1}{c|}{0.6858}  & \multicolumn{1}{c|}{23.49}   & 0.6516  & \multicolumn{1}{c|}{23.27}   & \multicolumn{1}{c|}{0.7117}  & \multicolumn{1}{c|}{22.75}   & 0.7157  \\ \hline
SRCNN~\cite{Dong2016Image}(+Bicubic)     & \multicolumn{1}{c|}{26.04}   & \multicolumn{1}{c|}{0.7005}  & \multicolumn{1}{c|}{23.98}   & 0.6654  & \multicolumn{1}{c|}{23.92}   & \multicolumn{1}{c|}{0.7246}  & \multicolumn{1}{c|}{23.47}   & 0.7295  \\ \hline
EDSR~\cite{lim2017enhanced}(+Bicubic)      & \multicolumn{1}{c|}{26.53}   & \multicolumn{1}{c|}{0.7192}  & \multicolumn{1}{c|}{24.91}   & 0.6916  & \multicolumn{1}{c|}{24.21}   & \multicolumn{1}{c|}{0.7323}  & \multicolumn{1}{c|}{23.98}   & 0.7452  \\ \hline
D-DBPN~\cite{haris2018ddbpn}(+Bicubic)    & \multicolumn{1}{c|}{26.59}   & \multicolumn{1}{c|}{0.7139}  & \multicolumn{1}{c|}{24.63}   & 0.6836  & \multicolumn{1}{c|}{24.39}   & \multicolumn{1}{c|}{0.7308}  & \multicolumn{1}{c|}{24.02}   & 0.7401  \\ \hline
RCAN~\cite{zhang2018image}(+Bicubic)      & \multicolumn{1}{c|}{26.70}   & \multicolumn{1}{c|}{0.7191}  & \multicolumn{1}{c|}{24.81}   & 0.6901  & \multicolumn{1}{c|}{24.40}   & \multicolumn{1}{c|}{0.7348}  & \multicolumn{1}{c|}{24.08}   & 0.7452  \\ \hline
360-SS~\cite{ozcinar2019super}(+Bicubic)    & \multicolumn{1}{c|}{23.28}   & \multicolumn{1}{c|}{0.6528}  & \multicolumn{1}{c|}{21.95}   & 0.6205  & \multicolumn{1}{c|}{22.00}   & \multicolumn{1}{c|}{0.6957}  & \multicolumn{1}{c|}{21.61}   & 0.6962  \\ \hline
LAU-Net~\cite{deng2021lau}(+Bicubic)   & \multicolumn{1}{c|}{26.39}   & \multicolumn{1}{c|}{0.7197}  & \multicolumn{1}{c|}{24.72}   & {0.6943}  & \multicolumn{1}{c|}{24.33}   & \multicolumn{1}{c|}{0.7346}  & \multicolumn{1}{c|}{24.97}   & {0.7727}  \\ \hline\hline
SphereSR(w/o SLIIF)(+Bicubic)           & \multicolumn{1}{c|}{26.66}   & \multicolumn{1}{c|}{0.7176}  & \multicolumn{1}{c|}{24.83}   & {0.6930}  & \multicolumn{1}{c|}{24.32}   & \multicolumn{1}{c|}{0.7345}  & \multicolumn{1}{c|}{25.00}   & {0.7477}  \\ %\hline
\hline
SphereSR(Ours)          & \multicolumn{1}{c|}{\textbf{26.76}}   & \multicolumn{1}{c|}{\textbf{0.7208}}  & \multicolumn{1}{c|}{\textbf{24.97}}   & {\textbf{0.6962}}  & \multicolumn{1}{c|}{{\textbf{24.46}}}   & \multicolumn{1}{c|}{{\textbf{0.7393}}}  & \multicolumn{1}{c|}{{\textbf{25.14}}}   & {\textbf{0.7780}}  \\ %\hline
\hline
\end{tabular}
\label{arbitrary}
\end{table*}

\begin{table*}[t!]
\small
\renewcommand{\tabcolsep}{3pt}
\renewcommand{\arraystretch}{0.9}
\centering
\caption{Ablation studies on ERP SR on ODI-SR and SUN360 Panorama Dataset for both $\times$8 and $\times$16 SR.
% Results are evaluated by WS-PSNR and WS-SSIM.
}
\vspace{-5pt}
\begin{tabular}{c|ccc|cccc|cccc}
\hline
Scale & \multicolumn{3}{c|}{Component}                                                                                                                                             & \multicolumn{4}{c|}{x8}                                                                              & \multicolumn{4}{c}{x16}                                                                             \\ \hline
\multirow{2}{*}{Model} & \multicolumn{1}{c|}{\multirow{2}{*}{GA-Conv}} & \multicolumn{1}{c|}{\multirow{2}{*}{SLIIF}} & \multirow{2}{*}{\begin{tabular}[c]{@{}c@{}}Feature\\      loss\end{tabular}} & \multicolumn{2}{c|}{ODI-SR}                                 & \multicolumn{2}{c|}{SUN 360 Panorama}  & \multicolumn{2}{c|}{ODI-SR}                                 & \multicolumn{2}{c}{SUN 360 Panorama}  \\ \cline{5-12} 
                       & \multicolumn{1}{c|}{}                       & \multicolumn{1}{c|}{}                         &                                                                              & \multicolumn{1}{c|}{WS-PSNR} & \multicolumn{1}{c|}{WS-SSIM} & \multicolumn{1}{c|}{WS-PSNR} & WS-SSIM & \multicolumn{1}{c|}{WS-PSNR} & \multicolumn{1}{c|}{WS-SSIM} & \multicolumn{1}{c|}{WS-PSNR} & WS-SSIM \\ \hline
1                      & \multicolumn{1}{c|}{x}                      & \multicolumn{1}{c|}{
\checkmark}                        & x                                                                            & \multicolumn{1}{c|}{24.20}   & \multicolumn{1}{c|}{0.6688}  & \multicolumn{1}{c|}{23.98}   & 0.6719  & \multicolumn{1}{c|}{22.44}   & \multicolumn{1}{c|}{0.6341}  & \multicolumn{1}{c|}{21.92}   & 0.6318  \\ \hline
2                      & \multicolumn{1}{c|}{\checkmark}                      & \multicolumn{1}{c|}{x}                        & x                                                                            & \multicolumn{1}{c|}{24.31}   & \multicolumn{1}{c|}{0.6731}  & \multicolumn{1}{c|}{24.07}   & 0.6749  & \multicolumn{1}{c|}{22.47}   & \multicolumn{1}{c|}{0.6335}  & \multicolumn{1}{c|}{21.92}   & 0.6294  \\ \hline
3                      & \multicolumn{1}{c|}{\checkmark}                      & \multicolumn{1}{c|}{\checkmark}                        & x                                                                            & \multicolumn{1}{c|}{24.34}   & \multicolumn{1}{c|}{0.6765}  & \multicolumn{1}{c|}{24.10}   & 0.6816  & \multicolumn{1}{c|}{22.47}   & \multicolumn{1}{c|}{0.6364}  & \multicolumn{1}{c|}{21.93}   & 0.6336  \\ \hline
4                      & \multicolumn{1}{c|}{\checkmark}                      & \multicolumn{1}{c|}{\checkmark}                        & \checkmark                                                                            & \multicolumn{1}{c|}{24.37}   & \multicolumn{1}{c|}{0.6777}  & \multicolumn{1}{c|}{24.17}   & 0.6820  & \multicolumn{1}{c|}{22.51}   & \multicolumn{1}{c|}{0.6370}  & \multicolumn{1}{c|}{21.95}   & 0.6342  \\ \hline
\end{tabular}
\label{ablation}
\end{table*}

\begin{table}[]
\centering
\small
\caption{Comparisons of activation memory between SpherePHD and our data structure. The network architectures of SpherePhD and ours have the same number of convolution layers (16) and hidden feature dimension (32).}
\vspace{-5pt}
\begin{tabular}{c|c|c|c|c}
\hline
Level               & 4   & 5    & 6    & 7     \\ \hline\hline
SpherePHD(MB)       & 660 & 1896 & 6714 & 26032 \\ \hline
New Data Structure(MB) & \textbf{374} & \textbf{724}  & \textbf{2138} & \textbf{7450}  \\ \hline
\end{tabular}
\vspace{-5pt}
\label{memory}
\end{table}

\subsection{Ablation Study and Analysis}
%이번 세션에서, 우리는 GA-Conv, SLIIF, Feature loss와 같은 우리가 제안한 서로 다른 각 모듈이 어느정도 영향력이 있는지를 살펴본다. 또한, GA-Conv에서 제안한 Data structure를 이용한 CNN implementation과 기존 SpherePHD[]의 CNN implementation의 activation memory 차이를 비교한다.

In this section, we study the effectiveness for each of our proposed modules, \eg, GA-Conv, SLIIF, and feature loss. In addition, we validate the memory load of CNN implementation between using the data structure in our GA-Conv and the existing SpherePHD~\cite{lee2020spherephd}.

\noindent \textbf{GA-Conv.  }
%우리는 GA-Conv의 유무에 따른 성능비교를 위해 Table 3에서 Model 1과 3의 결과를 비교한다. GA-Conv는 SLIIF에서 사용될 feature vector를 구하기 위한 feature extraction mudule에서 사용된다. GA-Conv를 사용하지 않는 경우는 SpherePHD[]에서 제안한 weight sharing, 즉 커널마다 180도 rotation을 주는 kernel weight sharing을 이용한다. 실험 결과,  scale factor x8, x16에서 모두 ODI-SR과 SUN360 Panorama dataset에서 성능향상이 있었다. 
We compare the results of Models 1 and 3 in Table~\ref{ablation} when adding or removing GA-Conv. GA-Conv is used in the feature extraction module to obtain the feature vector to be used in SLIIF. If GA-Conv is not used, the kernel weight sharing proposed by SpherePHD~\cite{lee2020spherephd}, which gives 180 degrees rotation per kernel, is used. Table~\ref{ablation} shows that using GA-Conv improves PSNR score by $0.14$dB and $0.12$dB in ODI-SR and SUN360 Panorama datasets, respectively, for $\times$8 SR.

\vspace{2pt}
\noindent \textbf{SLIIF.  }
%우리의 모델은 SLIIF를 이용하여 sphere상에 존재하는 feature vectors를 통해 ERP projection type에 대한 SR 결과를 predict한다. ~\cite{shi2016real}에서 언급한 방법과 동일하게, 우리는 SLIIF의 유무에 따른 성능비교를 위해, SLIIF없이 Icosahedral meshes상에서 super-resolution이 가능한 pixel-shuffle algorithm을 구현했다. 그리고 이 pixel-shuffle을 이용하여, 마지막 feature map을 scale factor 배수만큼 subdivision한 후, bicubic interpolation을 통해 최종 ERP output을 도출했다. Table 3의 Model 2와 3에서 볼 수 있듯이, icosahedron을 subdivision하는 pixel-shuffle 방식보다 continuous representation을 이용하는 SLIIF 방식이 더 높은 성능을 가지는 것을 확인할 수 있다. 
SphereSR uses SLIIF to present SR results for the ERP projection type through feature vectors presented on the sphere. Same as the method mentioned in ~\cite{shi2016real}, we implement a pixel-shuffle algorithm capable of performing SR on the Icosahedron without SLIIF for performance comparison.
% according to the presence or absence of SLIF. 
When using the pixel-shuffle, the last feature map was subdivided by scale factor multiple, and then the final ERP output was derived using the bicubic interpolation. As can be seen in Models 2 and 3 of Table~\ref{ablation}, using SLIIF for continuous image presentation achieves higher performance (24.34dB vs. 24.31dB) than the pixel-shuffle method of subdivision of icosahedron for $\times$8 SR.

\noindent \textbf{Feature loss.  }
%우리는 다른 projection type으로부터 생성된 feature를 이용하여 feature masking을 통해 중요한 영역에 대한 feature similarity를 구할 수 있는 feature loss를 제안한다. 본 loss의 유무에 따른 성능차이를 확인하기 위해서는, Table 3의 모델 3과 4를 비교해야 한다. 성능 비교 결과, x8, x16 scale에 대한 SR결과에서 모든 metric에 대하여 성능향상이 있는 것을 확인할 수 있다.
We propose a feature loss that measures the feature similarity of the crucial areas through feature masking using features generated from other projection types. To confirm the effectiveness of feature loss, we compare the SR performing when adding and removing this loss. Models 3 and 4 in Table 3 show the ablation results. Performance comparison indicates a performance improvement for all metrics in $\times$8 and $\times$16 SR.

\noindent \textbf{Data Representation Efficiency.} 
%우리는 GA-Conv 파트에서 기존 SpherePHD의 CNN implementation 방법이 효율적이지 않다는 것을 지적하고, 이러한 문제를 해결할 수 있는 새로운 data structure를 제안했다. 우리는 새로운 data structure가 얼마나 효율적으로 변했는지 확인하기 위해서, 간단한 CNN model을 구현한 후 Activation memory를 비교하는 실험을 진행했다. CNN model은 간단히 Convolution layer를 쌓아올린 구조이고, convolution 수는 16, hidden feature dimension은 32로 설정하였다.
%실험 결과, input level 4부터 7까지 모든 level에 대해서 new data structure가 더 낮은 activation memory를 가진다는 것을 확인하였다. 또한, input level이 증가할수록, SpherePHD 대비 New data structure memory 사용 비율이 감소한다는 것을 알 수 있었다. 본 결과를 토대로, 새롭게 제시한 data structure가 기존 spherePHD와 비교했을 때 더 메모리적인 면에서 효율적이고, input resolution이 증가할수록 효율성이 더 높아진다는 것을 알 수 있다.
In Sec.~\ref{ds_ks}, we point out that the CNN imposition method of SpherePHD is not efficient for SR. We thus propose a new data structure to tackle the problem. To see how efficiently the new data structure is, we implement a simple CNN model and then conduct an experiment comparing the activation memory. The CNN model is simply a structure in which a convolution layer is stacked, the number of convolution layers is set to 16, and the hidden feature dimension is set to 32.
Table.~\ref{memory} shows the experiments from input level 4 to 7. As can be verified, the new data structure in GA-Conv has a much lower activation memory. In addition, it is found that, as the input level increased, the ratio of using new data structure memory to SpherePHD decreased. Based on the result, it can be seen that the proposed data structure is more efficient in terms of memory compared to SpherePHD. Moreover, the efficiency increases as the input resolution increases.

\makeatletter
\newcommand{\thickhline}{%
    \noalign {\ifnum 0=`}\fi \hrule height 1pt
    \futurelet \reserved@a \@xhline
}
\newcolumntype{"}{@{\hskip\tabcolsep\vrule width 1pt\hskip\tabcolsep}}
\makeatother

%%%%%%%%%%%%%%%%%% Limitation and Future Work %%%%%%%%%%%%%%%%%%
% \section{Limitation and Future work}
%우리는 ODI를 input으로하여 유의미한 feature map을 추출하기 위해, 효율적인 data structure 및 새로운 convolution method(GA-Conv)를 찾는데 집중하였다. 그러나, 유의미한 feature map을 얻는 방법은 우리가 앞서 언급했던 내용 이외에도 network architecture를 수정하는 방법을 통해서도 이뤄질 수 있다. perspective image와 비교했을 때 ODI의 특수한 특성을 파악하여, 이러한 특성을 잘 반영할 수 있는 네트워크 architecture를 찾을 수 있다면, 우리가 구현해 둔 SLIIF를 통해 더 훌륭한 ODI-SR 결과를 얻을 수 있을 것이다.
% To extract meaningful feature maps with an ODI input, we focus on finding an efficient data structure and a kernel weight sharing method to better represent data. However, this be done by modifying the network architecture in addition to what we have mentioned in Sec.~\ref{ds_ks}. If we can identify the special characteristics of the ODI compared to perspective images and find a network architecture that can reflect these characteristics well, we can achieve better SR results via the proposed SLIIF.
%%%%%%%%%%%%%%%%%% Conclusion %%%%%%%%%%%%%%%%%%
\vspace{-3pt}
\section{Conclusion}
% \vspace{-3pt}
In this paper, we proposed a novel framework, called SphereSR, to generate a continuous spherical image representation from an LR 360$^{\circ}$ image. SphereSR predicts the RGB values at the given spherical coordinates of an HR image corresponding to an arbitrary project type. We first proposed geometry-aligned convolution to represent spherical data efficiently and then proposed SLIIF to extract RGB values from the spherical coordinates. As such, SphereSR flexibly reconstructed an HR image with an arbitrary projection type and SR scale factors. Experiments on various benchmark datasets showed that our method significantly surpasses existing methods.

\noindent\textbf{Limitation and Future Work}. We focused on finding an efficient data structure and kernel weight sharing method to extract meaningful features with the ODI input based on GA-Conv (Sec.~\ref{ds_ks}). %However, this be done by improving the network architecture . 
Future studies will therefore need to improve the network architecture using the properties of ODIs compared to perspective images, and then we can achieve better SR results via SLIIF.

%%%%%%%%% REFERENCES
\clearpage
{\small
\bibliographystyle{ieee_fullname}
\bibliography{egbib}
}

%\clearpage
%\input{supp}

\end{document}

\title{SphereSR: 360$^{\circ}$ Image Super-Resolution with Arbitrary Projection via Continuous Spherical Image Representation\\
--Supplementary Material--}

\author{Youngho Yoon, Inchul Chung, Lin Wang, and Kuk-Jin Yoon\\
Visual Intelligence Lab., KAIST, Korea\\
{\tt\small \{dudgh1732,inchul1221,wanglin,kjyoon\}@kaist.ac.kr }
%\and
}
\maketitle

\begin{abstract}
Due to the lack of space in the main  paper, we provide more details of the proposed methods and experimental results in the supplementary material. 
Specifically, in Sec.1, we provide more details of the feature extraction module for spherical images. Sec.2 explains explicitly the proposed sphere-oriented cell decoding in SLIIF module. 
Lastly, Sec.3 presents additional results about experiments and ablation study. 
\end{abstract}

%%%%%%%%% BODY TEXT
\section{More Details of Feature Extraction for Spherical Images}
\label{sec:feature}

%-------------------------------------------------------------------------
\subsection{Data structure}

\begin{figure}[b!]
  \centering
  \includegraphics[width=.7\linewidth]{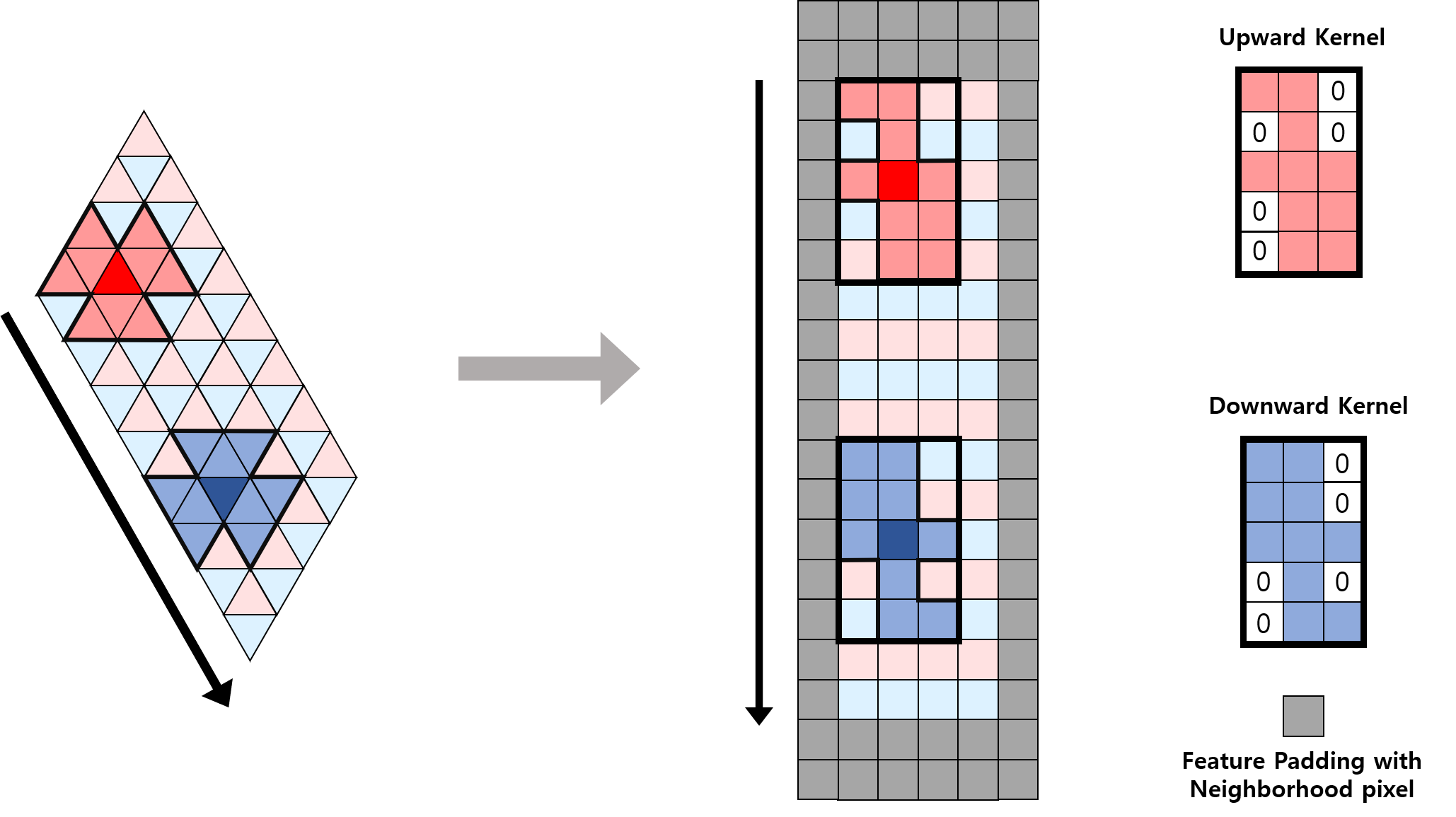}
  \vspace{-5pt}
   \caption{Proposed new data structure.}
   \label{fig:supp_data}
  \vspace{-10pt}
\end{figure}

%우리는 data structure를 변경함으로써 up/downward kernel의 구현을 5 by 3 convolution으로 가능하게 만들었다. 이때, 5 by 3 convolution kernel의 pixel 중 실제 kernel에 포함되지 않는 픽셀 5개에 대해서는 항상 weight 0이 곱해지며, weight update가 이뤄지지 않는다. 
%우리는 또한 spherical data structure의 연속성을 잃지 않기 위해, convolution 직전에 항상 feature padding with neighborhood pixel을 해준다. padding size는 kernel size를 고려하여 2 by 1 padding을 해준다.
%마지막으로, 홀수번째 row 및 짝수번째 row에 대해서 convolution되는 kernel이 다르기 때문에, 각 row별로 해당 kernel에 대한 convolution이 이뤄질 수 있도록 row axis에 대한 convolution stride를 2로 해준다. 각각의 kernel에 대해 convolution을 마친 후, feature concatenation을 해준다.

As shown in Fig.~\ref{fig:supp_data}, we make the implementation of up/downward kernel possible with 5$\times$3 convolution by using the proposed new data structure. At this time, among the pixels of the5$\times$3 convolution kernel, 5 pixels not included in the actual kernel are always multiplied by weight 0, and the weight update is not performed.

We also always do feature padding with neighborhood pixels right before convolution to avoid losing the continuity of the spherical data structure. The padding size is 2 $\times$ 1 padding considering the kernel size.

Finally, as the convolutional kernels for odd-numbered and even-numbered rows are different, the convolution stride for the row axis is set to 2 such that convolution for the kernel can be performed for each row. After convolution for each kernel, two feature maps are concatenated according to the order of rows.
%-------------------------------------------------------------------------
\subsection{Kernel Weight Sharing}

\begin{figure}[h!]
  \centering
  \includegraphics[width=.62\linewidth]{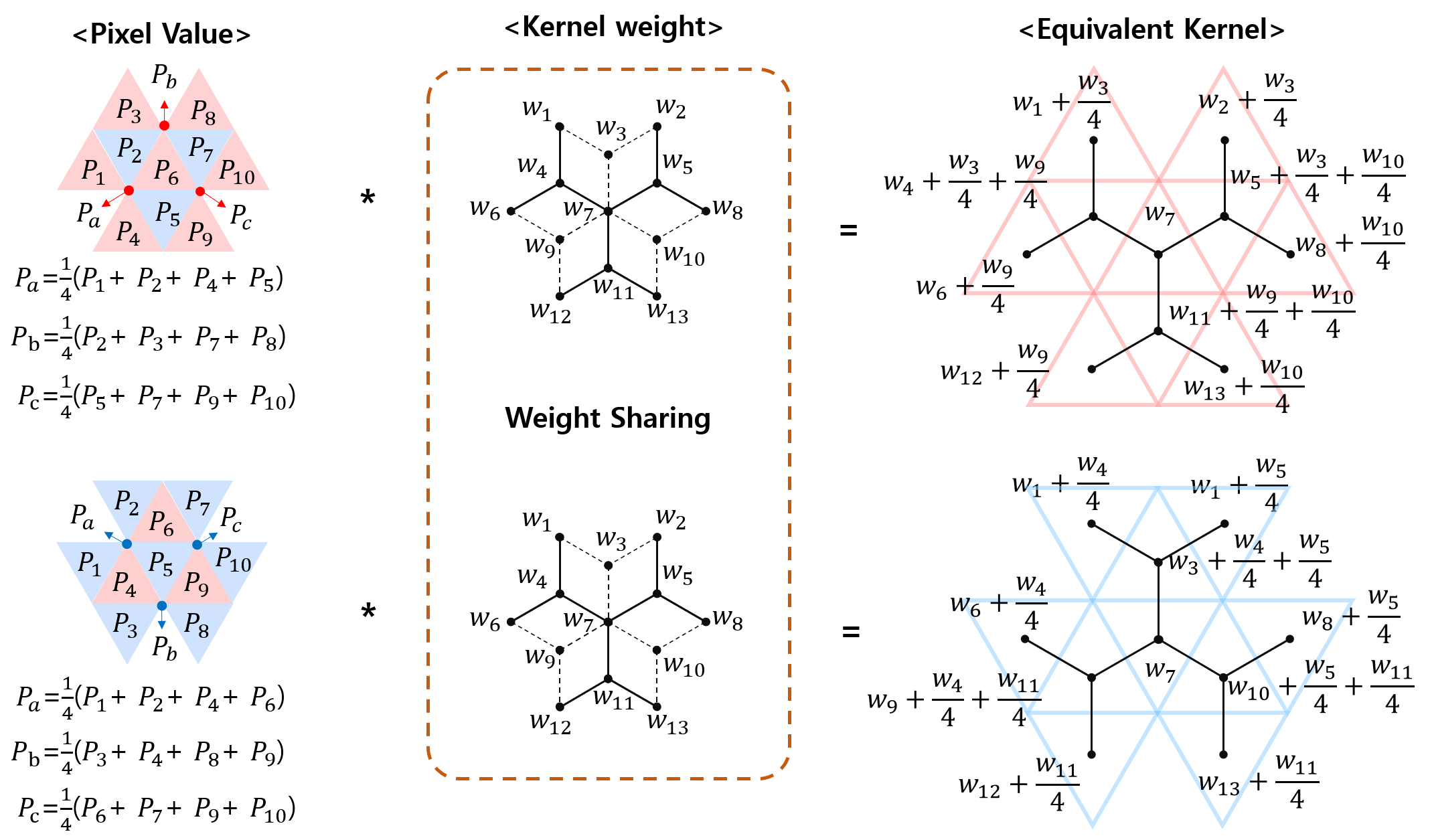}
  \vspace{-5pt}
   \caption{Geometry-Aligned Convolution(GA-Conv) and equivalent kernel.}
   \label{fig:supp_kernel}
  \vspace{-10pt}
\end{figure}

% 이번 파트에서, 우리는 GA-Conv(Geometry-Aligned Convolution)을 위한 kernel weight sharing scheme의 유도과정에 대해서 자세히 설명한다. 그림에서 볼 수 있듯이, 우리는 pixel combination의 shape을 동일하게 만들어주기 위해 center pixel의 세 vertices를 imaginary pixel $P_a,P_b,P_c$로 설정한다. 이때, 각 imaginary pixel은 주변의 네 픽셀의 중점이라 볼 수 있다. 따라서 imaginary pixel $P_a,P_b,P_c$의 value는 그림에 적힌 수식을 통해 네 픽셀의 mean value로 설정한다. 
%우리는 이러한 설정 시 implementation을 imaginary pixel을 실제로 만들지 않고, weight sharing을 활용할 수 있는 방법을 제시한다. 그림에서 볼 수 있듯이, pixel value와 kernel weight의 pixel-wise multiplication, 즉 convolution을 수행한 값은 equivalent kernel을 이용하여 convolution한 값과 동일하다는 것을 쉽게 알 수 있다. 따라서, GA-Conv를 사용하기 위해 적절한 weight sharing scheme을 사용하였고, 이를 통해 이전에 SpherePHD에서 이뤄졌던 kernel rotation 없이 더욱 정교한 kernel weight sharing이 가능해진다.

In this subsection, we describe in detail the derivation of the kernel weight sharing scheme for GA-Conv (Geometry-Aligned Convolution). As shown in Fig.~\ref{fig:supp_kernel}, we set the three vertices of the center pixel to the imaginary pixels $P_a, P_b, P_c$ to make the shape of the pixel combination the same. At this time, each pixel can be regarded as the midpoint of the surrounding four pixels. Therefore, the values of pixels $P_a, P_b, P_c$ are set as the mean value of four pixels through the formulas in the Fig.~\ref{fig:supp_kernel}.
We present a way to utilize weight sharing without actually creating a pixel when implementing this setting. As can be seen from Fig.~\ref{fig:supp_kernel}, the convolution with pixel values and kernel weights is the same as the value convolutional using the equivalent kernel. Therefore, an appropriate weight sharing scheme is used to use GA-Conv, and through this, more sophisticated kernel weight sharing is possible without kernel rotation previously performed in SpherePHD.

%------------------------------------------------------------------------
\section{Sphere-oriented Cell Decoding}
\label{sec:formatting}

In this section, we explain more details about sphere-oriented cell-decoding method.
A rectangle $\overrightarrow{\Delta X},\overrightarrow{\Delta Y}$ on the projected plane makes parallelogram $\overrightarrow{\Delta x},\overrightarrow{\Delta y}$ on the unit sphere. To define a cell decoding according to point $p$ on unit sphere, we initially define a vector $(\overrightarrow{n_1},\overrightarrow{n_2})$. Let $o = center\;point(r=0)$, $p = (\theta,\phi)$ s.t. point $p$ we want to get RGB and $v_x=(\theta_v,\phi_v)$s.t. one of vertices 1, 2, 3. Then, 
\begin{equation}
  \overrightarrow{n_1}=\overrightarrow{{v_x p}}=(\theta - \theta_v)\hat{\theta}+(\phi-\phi_v)\sin{\theta}\hat{\phi}
  \label{n1}
\end{equation}
\begin{equation}
  \overrightarrow{n_2}=\overrightarrow{o p}\times\overrightarrow{n_1}=-(\phi-\phi_v)\sin{\theta}\hat{\theta}+(\theta - \theta_v)\hat{\phi}
  \label{n2}
\end{equation}
To make $(\overrightarrow{n_1},\overrightarrow{n_2})$ as unit vectors, the vectors are scaled as
$\overrightarrow{n_1} \leftarrow \overrightarrow{n_1}/ \parallel\overrightarrow{n_1}\parallel$ and $\overrightarrow{n_2} \leftarrow \overrightarrow{n_2}/ \parallel\overrightarrow{n_2}\parallel$.
By these equations, we can find the variables $\alpha_1, \beta_1, \alpha_2, \beta_2$ that satisfies the following equation,
\begin{equation}
  \begin{aligned}
  \begin{pmatrix} \hat{\theta} \\ \hat{\phi} \end{pmatrix} 
  &=
  \begin{pmatrix} \alpha_1 & \beta_1 \\ \alpha_2 & \beta_2 \end{pmatrix} 
  \begin{pmatrix}\overrightarrow{n_1} \\ \overrightarrow{n_2} \end{pmatrix}
    %\label{n2}
  \end{aligned} 
\label{eq:1}
\end{equation}
   %\\&= 
   %\begin{pmatrix} \theta_x - \theta_v & (\phi_x-\phi_v)\sin{\theta_x} \\ -(\phi_x-\phi_v)\sin{\theta_x} & \theta_x - \theta_v \end{pmatrix}^{-1} 
   %\begin{pmatrix}\overrightarrow{n_1} \\ \overrightarrow{n_2} \end{pmatrix}
   %&&\label{n2}  
%\end{equation}

%We derive the relation between $\overrightarrow{\Delta X},\overrightarrow{\Delta Y}$ and $\overrightarrow{\Delta x},\overrightarrow{\Delta y}$ to find a cell decoding value of $\overrightarrow{\Delta X},\overrightarrow{\Delta Y}$. We explain the derivation process for each projection type, ERP, Perspective and Fisheye.

\begin{figure}[t!]
  \centering
  \includegraphics[width=.65\linewidth]{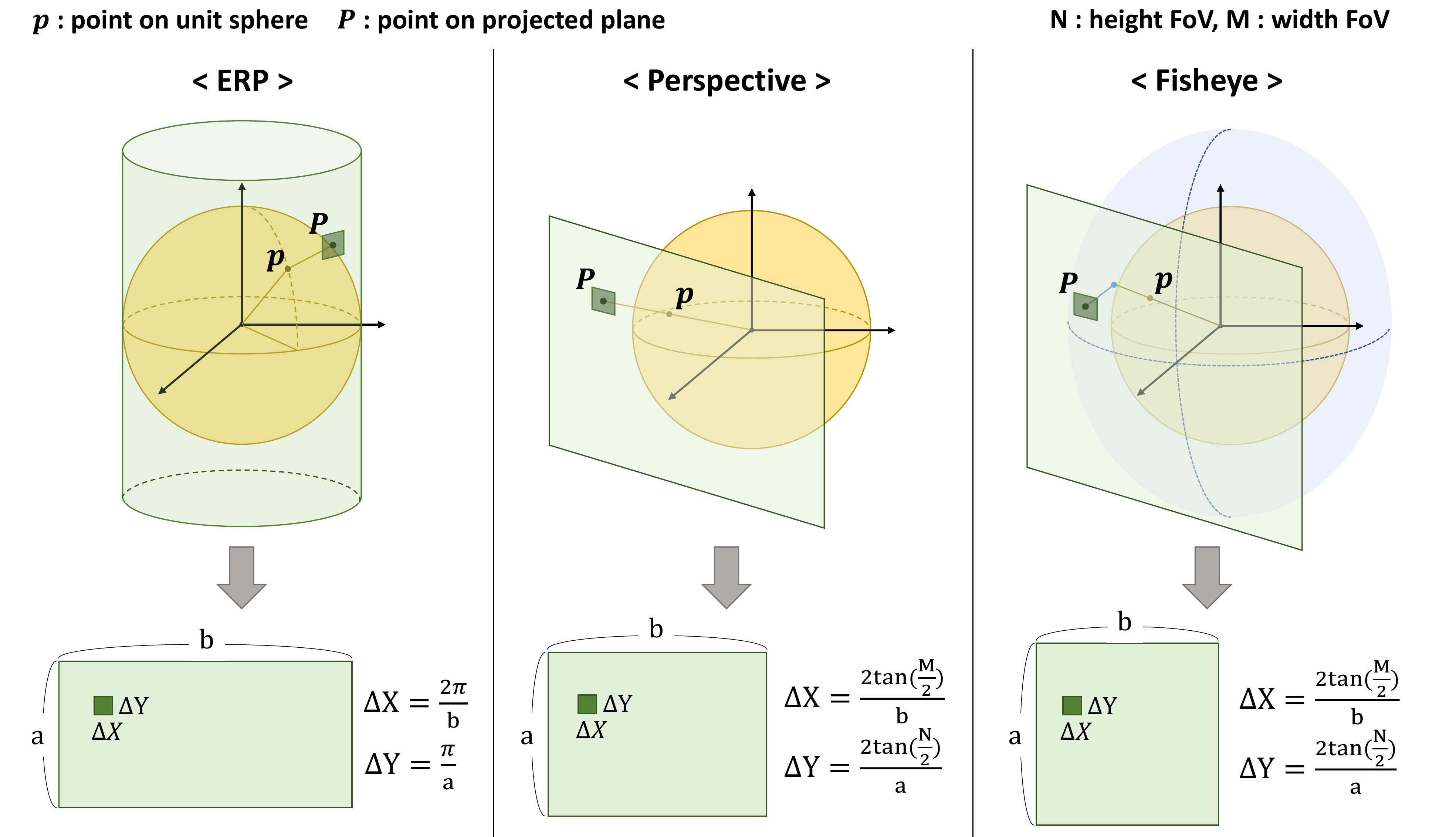}
  \vspace{-5pt}
   \caption{Visualization of projection process of each projection type.}
   \label{fig:supp_projection}
  \vspace{-10pt}
\end{figure}

To denote vectors $\overrightarrow{\Delta x},\overrightarrow{\Delta y}$ representing parallelogram sphere cells using $\overrightarrow{n_1}, \overrightarrow{n_2}$, we first denote using $\hat{\theta}, \hat{\phi}$ through the projection process of each projection type, as shown in Fig.~\ref{fig:supp_projection}. Therefore, we have to find as the equation,

\begin{equation}
  \begin{matrix}
    \overrightarrow{\Delta x} = \gamma_1 \hat{\theta} + \delta_1 \hat{\phi} \\
    \overrightarrow{\Delta y} = \gamma_2 \hat{\theta} + \delta_2 \hat{\phi}
  \end{matrix}
  \text{ } \Big) \Rightarrow 
  \begin{matrix}
    \text{decided by output of} \\
    \text{projection type}
  \end{matrix}
\label{eq:2}
\end{equation}

\begin{table*}[t!]
\caption{\label{tab:cell-decoding-tb}Derivation of $\vec{\Delta x}, \vec{\Delta y}$ for various projection types.}
\small
\renewcommand{\arraystretch}{0.7}
\renewcommand{\tabcolsep}{15pt}

\centering
\begin{tabular}{c|c}
\hline\hline
\multicolumn{1}{c}{ERP} & Perspective \\ \hline
\multicolumn{1}{l}{
$
  \begin{pmatrix}
    \overrightarrow{\Delta x} \\
    \overrightarrow{\Delta y}
  \end{pmatrix}
  =
  \begin{pmatrix}
    \Delta Y\sin{\theta} & 0 \\
    0 & \Delta X
  \end{pmatrix}
  \begin{pmatrix}
    \hat{\theta} \\
    \hat{\phi}
  \end{pmatrix}
$}  & 
{ $
  \begin{matrix}
  \begin{pmatrix}
    \overrightarrow{\Delta x} \\
    \overrightarrow{\Delta y}
  \end{pmatrix}
  =
  \begin{pmatrix}
    \alpha \Delta X \cos{\theta}\sin{\phi} & \alpha\Delta X \cos{\phi} \\
    -\alpha\Delta Y \sin{\theta} & 0
  \end{pmatrix}
  \begin{pmatrix}
    \hat{\theta} \\
    \hat{\phi}
  \end{pmatrix} \\
  \end{matrix}
    $}  \\ \hline\hline

\multicolumn{2}{c}{Fisheye} \\ \hline
\multicolumn{2}{l}{

$
\small
\renewcommand{\arraystretch}{0.9}
\renewcommand{\tabcolsep}{9pt}
  \begin{matrix}
  \begin{pmatrix}
    \overrightarrow{\Delta x} \\
    \overrightarrow{\Delta y}
  \end{pmatrix}
  =
  \begin{pmatrix}
    \alpha \Delta X \cos{\theta}\sin{\phi} + \alpha\Delta X F_{X}(X,Y)\cos{\theta}\cos{\phi} &
    \alpha\Delta X \cos{\phi} - \alpha\Delta X F_{X}(X,Y)\sin{\phi} \\
    -\alpha\Delta Y \sin{\theta} + \alpha\Delta Y F_{Y}(X,Y)\cos{\theta}\cos{\phi} &
    - \alpha\Delta Y F_{Y}(X,Y)\sin{\phi}
  \end{pmatrix}
  \begin{pmatrix}
    \hat{\theta} \\
    \hat{\phi}
  \end{pmatrix}
  \end{matrix}
$ } \\ \hline\hline

\end{tabular}

\end{table*}

Let P be the point $(X,Y)$ where point p is projected onto the projected plane. As the derivation results, each projection can be organized as a table.~\ref{tab:cell-decoding-tb}. Fisheye projection type is the result of the projection process by setting the reflective surface $Z=F(X,Y)$ as mentioned in ~\cite{Scaramuzza2006a}.
Also, in case of Perspective projection,
\begin{equation}
\begin{aligned}
\begin{matrix}
 
  \alpha=\frac{1}{ \sqrt {\tan^2{\phi} + \frac{1}{\tan^2{\theta}\cos^2{\phi}}}},  
  \Delta X = \frac{2\tan{\frac{M}{2}}}{b},
  \Delta Y = \frac{2\tan{\frac{N}{2}}}{a}

\end{matrix}
\end{aligned}
\end{equation}
in case of Fisheye projection,
\begin{equation}
\begin{aligned}
  \begin{matrix}

  \alpha=\frac{1}{ \sqrt {X^2 + Y^2 + F_{X}(X,Y)^2}} ,  
  \Delta X = \frac{F(X',0)}{b} \text{ s.t. } \frac{X'}{F(X',0)}=\tan{\frac{M}{2}},
  \Delta Y = \frac{F(0,Y')}{a} \text{ s.t. } \frac{Y'}{F(0,Y')}=\tan{\frac{N}{2}}

  \end{matrix}
\end{aligned}
\end{equation}

Finally, we are able to represent $\overrightarrow{\Delta x},\overrightarrow{\Delta y}$ using two vectors $\overrightarrow{n_1}, \overrightarrow{n_2}$ by \cref{eq:1,eq:2}.
\begin{equation}
  \begin{aligned}
  \begin{pmatrix} \overrightarrow{\Delta x} \\ \overrightarrow{\Delta y} \end{pmatrix} 
  &= 
  \begin{pmatrix} \gamma_1 & \delta_1 \\ \gamma_2 & \delta_2 \end{pmatrix}
  \begin{pmatrix} \alpha_1 & \beta_1 \\ \alpha_2 & \beta_2 \end{pmatrix} 
  \begin{pmatrix}\overrightarrow{n_1} \\ \overrightarrow{n_2} \end{pmatrix}
    %\label{n2}
  \end{aligned} 
\label{eq:sum}
\end{equation}
We then find a  $\overrightarrow{\Delta x_{eq}}, \overrightarrow{\Delta y_{eq}}$ as mentioned in the main paper.

%우리는 또한 parallelogram sphere cell을 나타내는 벡터들 $\overrightarrow{\Delta x},\overrightarrow{\Delta y}$을 $\overrightarrow{n_1}, \overrightarrow{n_2}$로 나타내기 위해서 먼저 각 projection type의 projection process을 통해 $\hat{\theta}, \hat{\phi}$로 유도할 수 있다.

%unit sphere상의 point $(\theta,\phi)$를 p, 그리고 point p가 projected plane 상으로 투영된 point $(X,Y)$를 P라 하자. 유도과정을 통해 결과를 나타내면 각각의 projection에 대해 표와 같이 정리할 수 있다. Fisheye projection type은 reflective surface $Z=F(X,Y)$를 설정하여 projection process를 거친 결과이다~\cite{Scaramuzza2006a}. 

%------------------------------------------------------------------------
\section{Visualization}
\label{sec:formatting}
\subsection{Additional Qualitative Comparison.} 
\begin{figure}[p]
  \centering
  \includegraphics[width=.9\linewidth]{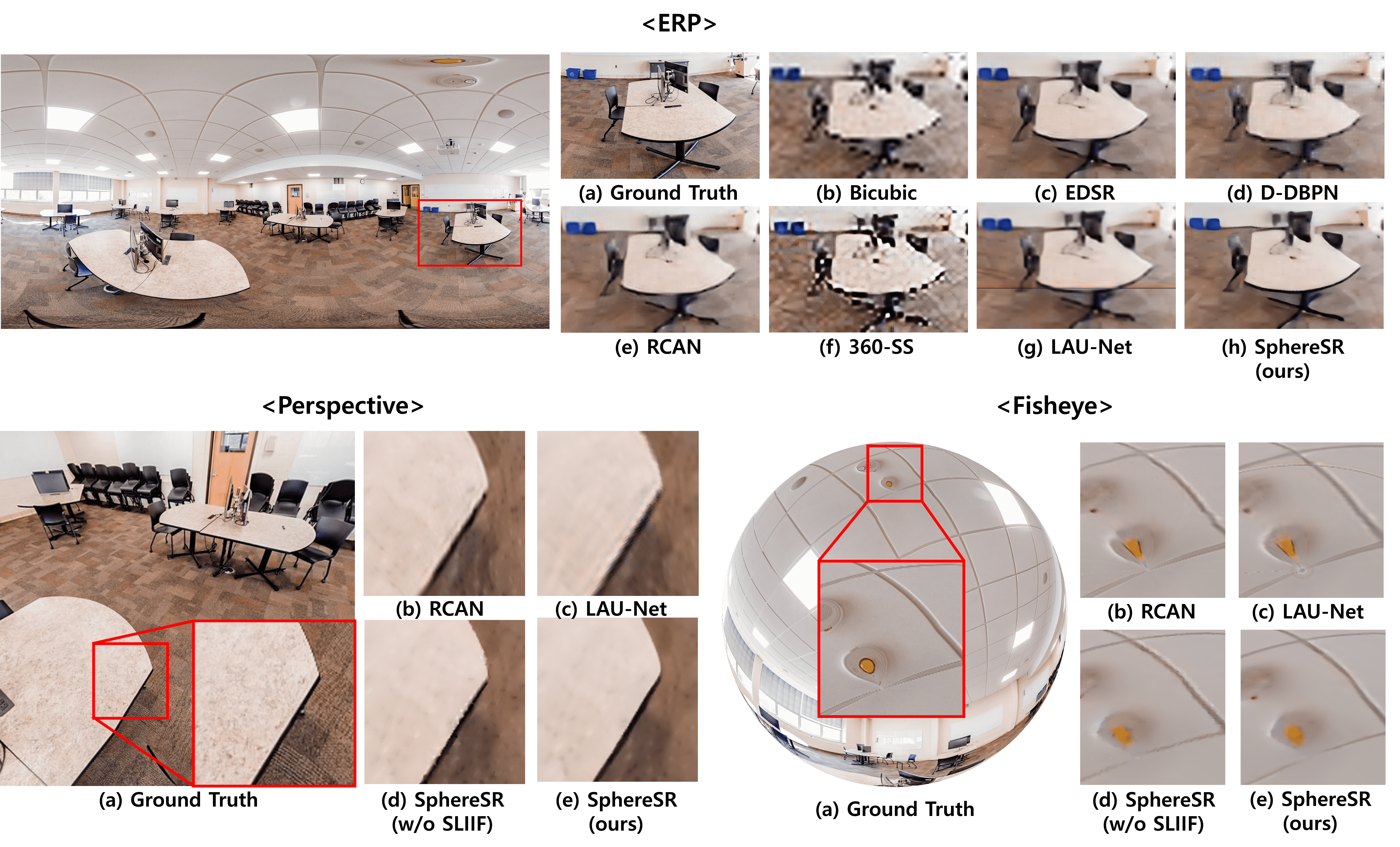}
  \vspace{-5pt}
   \caption{Visual comparisons of $\times$8 SR results of ERP, perspective, fisheye projection on SUN 360 Panorama dataset.}
   \label{fig:supp_comparison1}
  \vspace{-10pt}
\end{figure}

\begin{figure}[p]
  \centering
  \includegraphics[width=.9\linewidth]{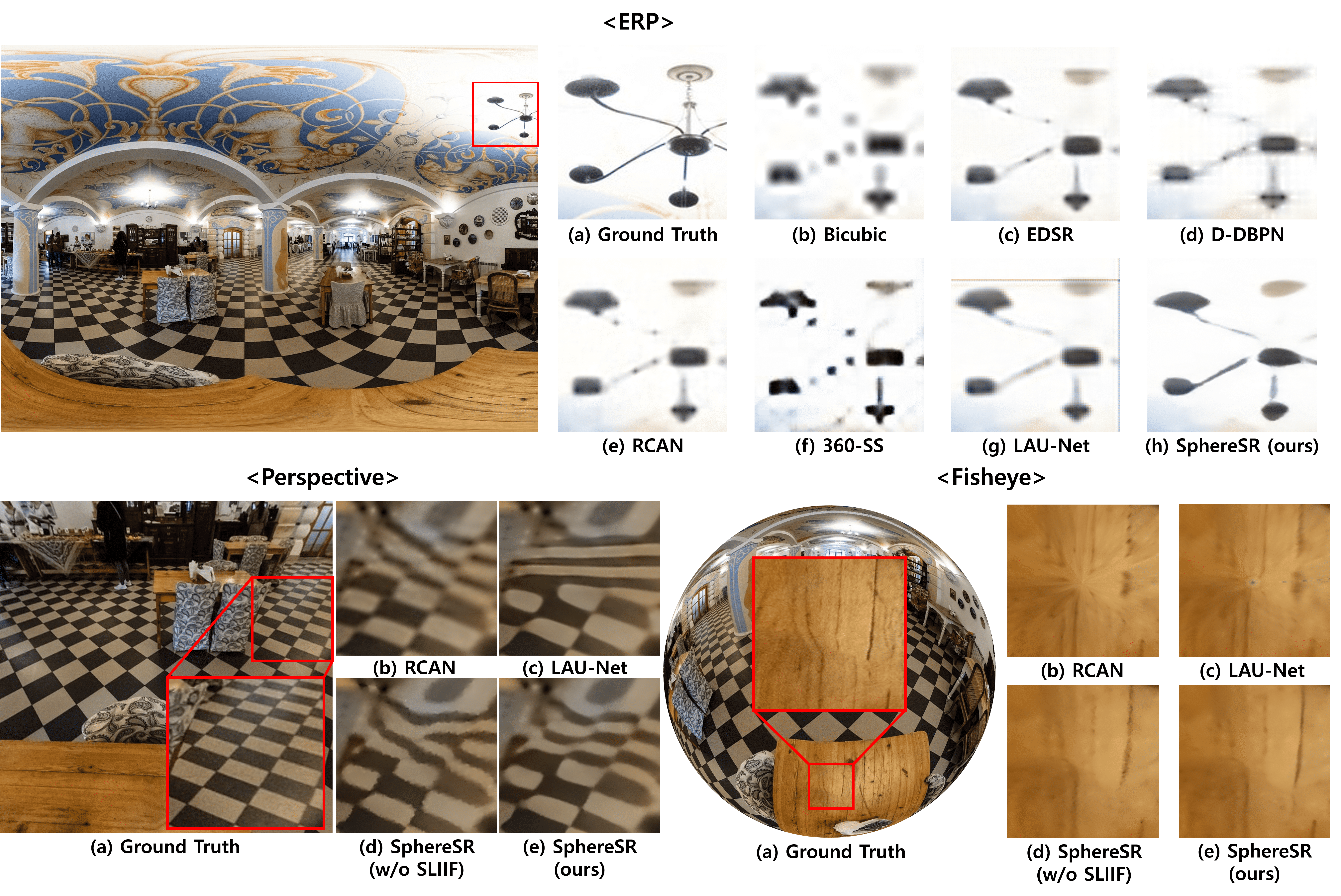}
  \vspace{-5pt}
   \caption{Visual comparisons of $\times$8 SR results of ERP, perspective, fisheye projection on ODI-SR dataset.}
   \label{fig:supp_comparison2}
  \vspace{-10pt}
\end{figure}

We additionally visualize the qualitative comparison with Ours and other methods on ODI-SR dataset and SUN360 Panorama dataset in Fig.~\ref{fig:supp_comparison1} and Fig.~\ref{fig:supp_comparison2}. We compare the results with various projection types, ERP, perspective and fisheye.

\subsection{Visualization of SR with Arbitrary Projection Types.}
\begin{figure}[p]
  \centering
  \includegraphics[width=.99\linewidth]{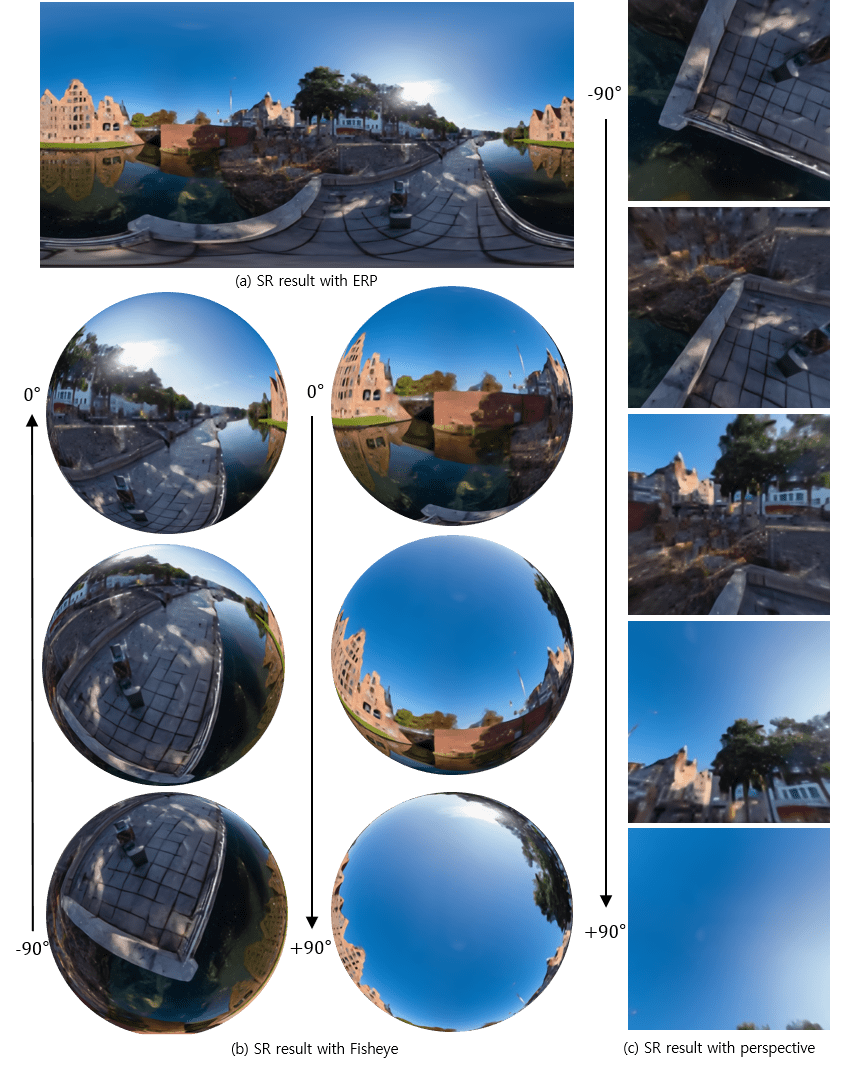}
  \vspace{-5pt}
   \caption{Visualization of $\times$8 SR results with various projection types on SUN360 Panorama dataset.}
   \label{fig:supp_various1}
  \vspace{-10pt}
\end{figure}

\begin{figure}[p]
  \centering
  \includegraphics[width=.99\linewidth]{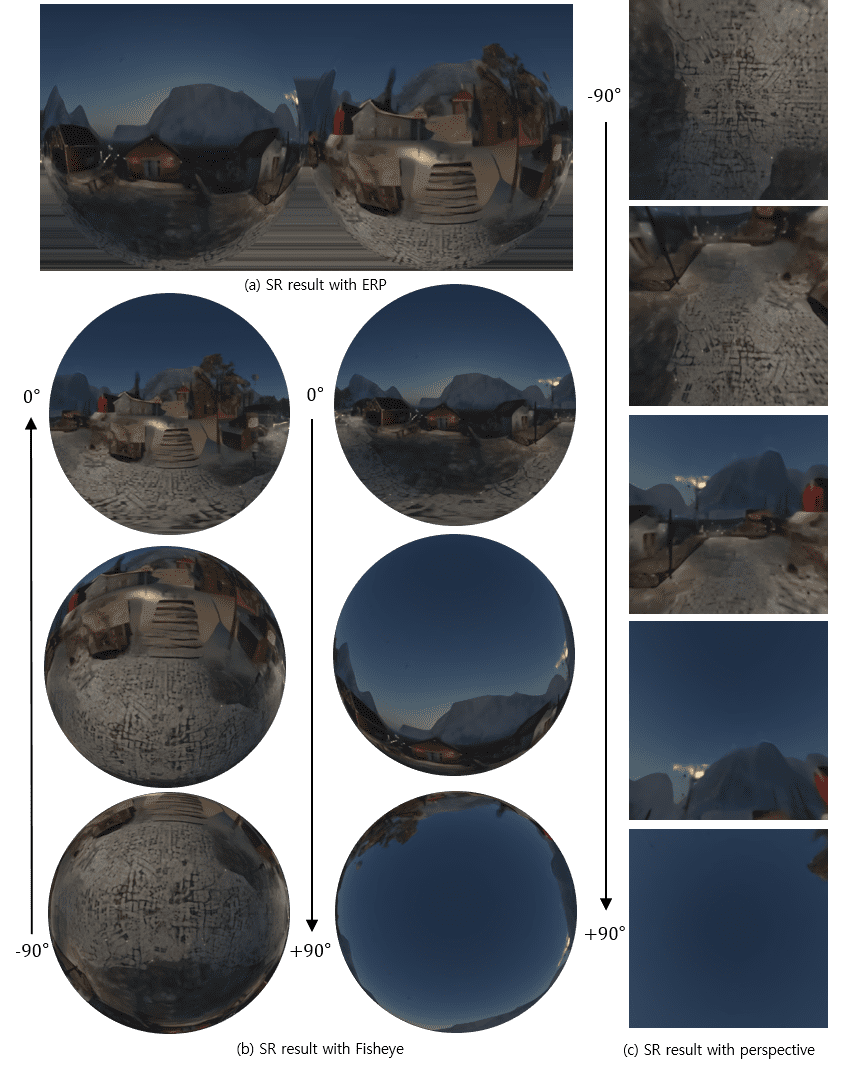}
  \vspace{-5pt}
   \caption{Visualization of $\times$8 SR results with various projection types on ODI-SR dataset.}
   \label{fig:supp_various2}
  \vspace{-10pt}
\end{figure}

We additionally visualize the SR results for arbitrary projection type from LR ERP images in Fig.~\ref{fig:supp_various1} and Fig.~\ref{fig:supp_various2}. We also visualize images for different directions to show that our model can predict SR results for arbitrary directions.

%이미지 한 장에 대해서 여러 projection type으로 가는 그림을 많이 넣어줘야 함. 되도록이면 임의의 projection type을 좀 더 다양하게 한다면 좋을 텐데 이 부분은 힘들 것 같다.
%All text must be in a two-column format.

%%%%%%%%% REFERENCES
{\small
\bibliographystyle{ieee_fullname}
\bibliography{egbib}
}